# An Integrated Optimization and Machine Learning Models to Predict the Admission Status of Emergency Patients

Abdulaziz Ahmed[a*], Omar Ashour[b], Haneen Ali[c], Mohammad Firouz[d]

[a]*Department of Health Services Administration, School of Health Professions, The University of Alabama at Birmingham, Birmingham, Alabama, USA*

[a]*Department of Industrial Engineering, The Pennsylvania State University, Erie, PA, USA*

[c]*Healthcare Services Administration Program, Auburn University, Auburn, AL, USA*

[d] *Department of Management, Information Systems & Quantitative Methods, Collat School of Business, University of Alabama at Birmingham, Birmingham, AL, USA*

---

[*] Corresponding author aahmed2@uab.edu
Address: 1720 University Blvd, Birmingham, AL 35294
Phone: +1 (205) 598-3531
Email addresses: A. Ahmed (aahmed2@uab.edu) O. Ashour (oma110@psu.edu), H. Ali (hba0007@auburn.edu , M. Firouzd (mfirouz@uab.edu).

**ABSTRACT**

This work proposes a framework for optimizing machine learning algorithms. The practicality of the framework is illustrated using an important case study from the healthcare domain, which is predicting the admission status of emergency department (ED) patients (e.g., admitted vs. discharged) using patient data at the time of triage. The proposed framework can mitigate the crowding problem by proactively planning the patient boarding process. A large retrospective dataset of patient records is obtained from the electronic health record database of all ED visits over three years from three major locations of a healthcare provider in the Midwest of the US. Three machine learning algorithms are proposed: T-XGB, T-ADAB, and T-MLP. T-XGB integrates extreme gradient boosting (XGB) and Tabu Search (TS), T-ADAB integrates Adaboost and TS, and T-MLP integrates multi-layer perceptron (MLP) and TS. The proposed algorithms are compared with the traditional algorithms: XGB, ADAB, and MLP, in which their parameters are tunned using grid search. The three proposed algorithms and the original ones are trained and tested using nine data groups that are obtained from different feature selection methods. In other words, 54 models are developed. Performance was evaluated using five measures: Area under the curve (AUC), sensitivity, specificity, F1, and accuracy. The results show that the newly proposed algorithms resulted in high AUC and outperformed the traditional algorithms. The T-ADAB performs the best among the newly developed algorithms. The AUC, sensitivity, specificity, F1, and accuracy of the best model are 95.4%, 99.3%, 91.4%, 95.2%, 97.2%, respectively.

## 1. INTRODUCTION

Emergency departments (EDs) are responsible for the majority of hospital admissions even though most ED visits result in a discharge (Moore et al., 2017). In 2017, there were around 139 million visits to the EDs in the U.S., 14.5 million (10.4%) led to hospital admissions, and 2 million led to admission to the critical care unit (Rui & Kang K, 2017). Due to the complexity and the wide array of complaints and injuries, EDs are usually overcrowded (Ashour & Kremer, 2016; Ashour & Kremer, 2013; Chonde et al., 2013). Overcrowding has been correlated to poor healthcare outcomes such as higher mortality rates, ambulance diversions, treatment delays, patients leaving without being seen, etc. (Moore et al., 2017; Arya et al., 2013; Sun et al., 2013; Araz et al., 2019).

Multiple approaches to alleviating the effect of crowding such as the use of fast-track approaches, triage, and lean six sigma have been proposed (Ashour & Kremer, 2016; Ashour & Kremer, 2013; Ben-Tovim et al., 2008; Chonde et al., 2013; Considine et al., 2008; Dickson et al., 2009; Holden, 2011; Kelly et al., 2007; King et al., 2006; Rodi et al., 2006). Triage is one of the important tools used to manage time effectively and improve ED operational performance (van der Vaart et al., 2011). Triage involves sorting incoming patients into groups according to their urgency level. It is usually done by a nurse based on patients' information such as demographics, chief complaints, and vital signs (Hong et al., 2018). Some patients arrive with life-threatening conditions and others can wait. Triage is a very important step that is used to facilitate patient flow and improve patients' safety and quality of care and as a result, reduce overcrowding at EDs (Ashour & Kremer, 2016; Ashour & Kremer, 2013; Chonde et al., 2013). Once the patient is triaged, s/he is examined by a healthcare worker, who provides the healthcare delivery and evaluates the patient's disposition. If the disposition decision was to admit the patient, the patient goes through the process of boarding which includes bed assignment and transportation (Lee et al., 2020). Unfortunately, the decisions made across different areas (e.g., ED and inpatient units) are not usually coordinated and that could lead to inefficient patient flow and delays in hospitals. In addition, the triage decision is subjective because it depends heavily on the nurse's knowledge and experience. The lack of coordination and inconsistencies in decisions present sources of variabilities that degrade the performance of the ED system and introduce problems such as overcrowding. Thus, accurate and consistent prediction models are needed to assess healthcare providers in making crucial decisions that improve patient outcomes.

One of the crucial factors to overcrowding are the delays occurring in the EDs due to the patient boarding process (Fatovich et al., 2005; Hoot & Aronsky, 2008; Lee et al., 2020; Pines et al., 2011; Pines & Bernstein, 2015). A study found a positive relationship between boarding delays and an intensive care patient's length of stay (LOS)(Chalfin et al., 2007). Improving the inpatient discharge times can reduce patient boarding delays significantly (Shi et al., 2016). Improvements such as the quick identification of the admission status during triage or proactively preparing the downstream resources could result in reducing the boarding times (Lee et al., 2020; Qiu et al., 2015; Peck et al., 2012). Thus, the availability of prediction models can help identify the admission status of incoming patients as well as the patients' mix which helps to manage downstream resources and reduce overcrowding at EDs by shortening boarding delays (Arya et al., 2013; Dugas et al., 2016). In the past few years, machine learning algorithms have been well improved and implemented in many applications such as preventive medicine (C.-S. Yu et al., 2020), hospital operations (Bacchi et al., 2020), and cancer detection (Saba, 2020). However, one of the most challenging problems with machine learning is that every algorithm has parameters and without optimizing these parameters, obtaining a high accuracy model becomes very difficult (Sarkar et al., 2019).

This study proposes a framework based on integrated optimization-machine learning algorithms to accurately predict two main ED disposition outcomes (discharge and admission). A metaheuristics optimization algorithm is utilized to optimize three machine learning algorithms. The admitted decision implies that an ED patient is hospitalized to an inpatient unit, while the discharged decision implies that the patient does not need hospitalization and is discharged from ED and sent home. The goal of the proposed framework is to early predict whether a patient needs to be admitted (i.e., hospitalized) or discharged once the patient arrives at an ED. This helps healthcare providers to coordinate with downstream units ahead of time to allow for bed assignment and transportation coordination and as a result, reduce boarding delays and eventually mitigate the ED crowding problem. The prediction models utilize basic and available triage data that is recorded initially in the ED patient visit. The framework includes three main phases: Data preprocessing, feature selection, and model development. Four feature selection algorithms are used:

decision tree (DT), random forest (RF), least absolute shrinkage and selection operator logistic regression (lasso-LR), and Chi-square (Chi-sq). The four feature selection methods are executed using multiple Scikit-learn functions, which results in various data group combinations. Each data group is then used to train and test six machine learning algorithms: T-XGB, T-ADAB, T-MLP, XGB, ADAB, and MLP, in which T-XGB, T-ADAB, T-MLP are newly proposed. The new algorithms are a result of the integration of Tabu search (TS) with three predictive algorithms: Extreme gradient boosting (XGB), Adaboost (ADAB), and multi-layer perceptron (MLP). The motivation of integrating Tabu search (TS) with the predictive algorithms is to achieve higher accuracy in the resulting models. Performance is evaluated using five measures: Area under the curve (AUC), sensitivity, specificity, F1, and accuracy. This work's contributions include:

- This work considers the hyperparameter fine-tuning problem in machine learning as an optimization problem and presents a comprehensive framework for integrating machine learning and metaheuristics to solve it.
- The proposed work shows how TS can be used to optimize three machine learning algorithms: XGB, ADAB, and MLP. Most of the parameters of the three algorithms are considered for optimization (e.g., five parameters for each algorithm).
- The proposed models are applied to imbalanced data (e.g., admission status of ED patients).
- The findings of this study are based on a large sample size dataset that is collected from different regions in the Midwest of the US. Therefore, the results are practical, generalizable, and more robust.
- The optimized best model will be used by the technology department at healthcare provider who provided us with the data as a decision tool, which will be used to improve patient flow and ultimately mitigate the ED crowding problem at different locations of the partner hospital who provided us with the data, especially in large metropolitan areas.

The paper is organized into the following sections: Section 2 provides a literature review of related prediction models of patient admission status at EDs as well as the use of metaheuristic approaches in machine learning. Section 3 describes the proposed research framework including a description of the data, feature selection algorithms, metaheuristics, prediction models, and performance measures. Section 4 provides the experimental, optimization, and results for the ED admission prediction. Finally, section 5 concludes by offering insights for future works.

## 2. RELATED WORK

### 2.1 Prediction Models in Emergency Departments (EDs)

Several studies have investigated ways to reduce boarding delays and their impact on overcrowding at EDs. For example, the impact of "early task initiation" such as proactively identifying the admission status or proactively preparing the downstream resources on reducing boarding times (Barak-Corren, Israelit, et al., 2017; Lee et al., 2020; Peck et al., 2012; Qiu et al., 2015). One way to proactively manage resources and reduce overcrowding at EDs is to predict the patient mix and use that information to manage ED resources as well as the downstream operations including hospital bed assignments and the need for emergency procedures (Arya et al., 2013; Dugas et al., 2016; Levin et al., 2018; Peck et al., 2012). Predictive modeling can be used to improve healthcare operations and efficiency (Moons et al., 2012; Obermeyer, Ziad & Emanuel, Ezekiel J., M.D., 2016; Peck et al., 2012). Chonde et al. (2013) developed and compared three models (e.g., artificial neural networks (ANNs), ordinal logistic regression (OLR), and naïve Bayesian networks (NBNs)) to predict the patient's emergency severity index (ESI) at EDs. ESI is a triage algorithm that organizes ED patients into 5 levels that reflect the severity of their symptoms (Tanabe et al., 2004). Golmohammadi (2016) implemented neural networks (NNs) and logistic regression models to identify the relationships among patients' characteristics such as age, radiology images, and the admission decision. Another study developed models to predict early readmissions to hospitals (Futoma et al., 2015). Graham et al. (2018) applied predictive algorithms including logistic regression, gradient boosting algorithms, and decision trees to predict admission status at EDs. Other researchers developed models for acute coronary syndrome or predict sepsis to help health systems identify terminal conditions while other prediction models were developed to help improve patient flow or hospital utilization at the system level (Haimovich et al., 2017; Horng et al., 2017; Y. Sun et al., 2011; Taylor et al., 2016; Weng et al., 2017).

Several studies have used patient's triage information, such as chief complaint, vital signs, age, gender, etc., to group patients and predict hospital admission decisions and/or improve resource utilization and patient flow (El-Darzi et al., 2009; Lucini et al., 2017; Lucke et al., 2018). In addition to triage information, other studies used the system and administrative information (Fine, et al., 2017; El-Darzi et al., 2009). The Glasgow Admission Prediction Score and the Sydney Triage to Admission Risk Tool are examples of formalized tools that are based on models built with the use of triage information Adding more information such as lab test results, and medications are given, and diagnoses tend to improve models' accuracy and predictive power. Some of this information can be extracted from the patient's previous visits and including this information could lead to more robust predictive models (Hong et al., 2018), however, this information usually is not available at the time of triage.

Many studies have used logistic regression and Naive Bayes modeling to forecast admission results (Israelit, et al., 2017; Peck et al., 2012; Leegon et al., 2005). Few studies have used complex algorithms modeling such as random forests, support vector machines (SVM), and artificial neural networks (NNs)(Leegon et al., 2006; Levin et al., 2018; Lucini et al., 2017). One recent study has used gradient boosting (XGB) and deep neural networks (DNN) to forecast admission at ED triage (Hong et al., 2018). They used the following features: Previous healthcare statistics, patient medical records, previous lab, and vital results, outpatient medications, past imaging counts, and demographic details such as insurance and employment status.

Despite the abundance of admission status prediction models in the literature, there are no widely adopted admission status prediction models in practice due to many reasons such as the requirement of specific patient data that are not available during the triage stage, and many of these models are built for a specific population or disease (Parker et al., 2019). Another reason that can be added is the tradeoff between building a simple model that has high accuracy. In other words, no scoring system has both simplicity and enough accuracy to be used in clinical settings (Cameron et al., 2018). In this work, a practical framework that is based on optimized prediction models is developed to determine the admission status of ED patients to help healthcare providers make informed decisions about the admission status of ED patients and as a result, better manage hospital resources, reduce delays in EDs, and improve care safety and quality.

## 2.1 Approaches Used for Optimizing Machine Learning

Several approaches have been proposed to optimize machine learning hyperparameters such as grid search (Bergstra et al., 2013) and random search (Bergstra & Bengio, 2012; Putatunda & Rama, 2019). In grid search, a search space is defined as a grid that contains hyperparameter values and when a model training is performed, every position in the grid is evaluated. However, the grid search approach is flawed because the number of times a model is evaluated grows exponentially as the number of parameters increases. While all possible combinations of different hyperparameters are tested, gird search is time-consuming, and finding the optimal value of a model hyperparameter is not guaranteed (Bergstra et al., 2011). In random search, a search space is defined as a bounded domain of hyperparameter values within which the hyperparameter values are randomly selected. However, the random search approach is flawed because it involves high variance and does not converge to a global optimum (Andradóttir, 2015). Aside from grid and random search approaches, an automated approach was proposed to optimize machine learning. Li et al. (2017) proposed a framework based on the Bayesian optimization method to optimize, convolutional neural networks (CNN), and support vector machines (SVM). Further, the parameters of XGB have been optimized using Bayesian optimization (Guo et al., 2019). However, the Bayesian optimization approach is flawed because the efficiency degrades as the number of parameters gets too large.

The body of literature on optimizing the hyperparameters of machine learning algorithms lacks approaches and methodologies that use efficient optimization techniques to determine the necessary hyperparameters. For example, Badrouchi et al. (2021) used XGB, MLP, K-nearest neighbor, and logistic regression to predict the survival of kidney transplants. They implemented a grid search technique (GridSearch Scikit-learn function) for parameter fine-tuning. The major difference between Badrouchi et al.'s (2021) work and this paper serves as the integration of optimization metaheuristic approaches with machine learning algorithms. A limited number of studies have utilized metaheuristic approaches to optimize machine learning algorithms. For example, particle swarm optimization (PSO) and genetic algorithm (GA) algorithms are the most used approaches. GA and PSO have been used to improve the hyperparameters of SVM (Pham & Triantaphyllou, 2011; Chou et al., 2014), and artificial neural networks (ANN) (Sarkar et al., 2019). In SVM, the optimized parameter is only one, which is gamma, while two parameters are considered for neural networks, which are the number of nodes, and learning rate. GA was also used to optimize the hyperparameters of XGB (Chen et al., 2020). The problem of GA and PSO is that they are population-based metaheuristics, which increases their computational cost. Simulated annealing (SA) algorithm has been used to determine the optimal value of one parameter for Deep Neural Network (DNN), which is the number of hidden layers (Tsai & Li, 2009). Bereta (2019) used TS to determine the optimal number of weak-learner of ADAB, while in this study, the parameters of both base-learner and meta-learner of ADAB are optimized. In addition, the domain applications are different and varied from public transportation systems (Tsai & Li, 2009), public-private partnership dispute (Chou et al., 2014), adverse occupational events (Sarkar et al., 2019), and face recognition (Bereta, 2019). Based on the reviewed studies, the literature gaps are:

- Most of the previous studies considered population-based metaheuristic algorithms for optimizing machine learning.
- Few parameters of machine learning are considered for optimization (e.g., Gamma in SVM or number of hidden neurons in neural networks).
- No studies have shown the performance of optimized machine learning algorithms (e.g., GA-SVM) when applied to imbalanced data, knowing that it is difficult to achieve high accuracy with imbalanced data (Badrouchi et al., 2021).

## 3. RESEARCH METHODOLOGY

This section presents the research methodology of this study including the proposed framework, feature selection, and model development.

### 3.1 Research Framework

The newly developed framework of this study is shown in Figure 1. Phase I describes the data collection and sources as well as the data preprocessing procedure (e.g., handling missing data, data scaling, etc). Phase II starts with visualizing the data to understand the input and output features. Then, feature selection is conducted to identify the most important features and avoid overfitting. Four feature selection algorithms are used: RF, DT, LASSO-LR, and Chi-sq, which are implemented via three Scikit-learn functions: SFM, RFE, and SKB. The SFM and RFE functions are utilized to implement LASSO-LR, RF, DT, while Chi-sq is executed using the SKB Scikit-learn function. Nine groups are obtained from the seven feature selection stage as follows: (1) Lasso_LR_SFM, (2) RF_SFM, (3) DT_SFM, (4) Chi-sq_SKB, (5) Lasso_LR_RFE, (6) RF_RFE, (7) DT_RFE, (8) voting group, and (9) all features in one group. The voting group includes the features that are selected by at least three selection algorithms out of the seven algorithms. The nine data groups obtained from the feature selection step are then used to develop six algorithms: T-XGB, T-ADAB, T-MLP, XGB, ADAB, and MLP. In short, 54 models are built ((7 data groups from phase II + one group that includes features from voting + one group that represents all the features) × 6 prediction algorithms = 54 models). In T-XGB, T-ADAB, T-MLP, hyperparameters are optimized using TS, while in XGB, ADAB, MLP are tunned using grid search. To overcome the data imbalance problem, the Synthetic Minority Oversampling Technique (SMOTE) is used for oversampling. The AUC was used as a performance measure when optimizing the three prediction algorithms. Next, five performance measures are used to evaluate the models with the best parameters: Accuracy, sensitivity, specificity, F1 measure, and AUC. The model with the best overall performance is selected.

### 3.2 Data Collection-Preprocessing

Retrospective data are obtained from one of the largest hospital systems in the Midwest in the U.S. It has more than 40 medical centers and 210 clinics located in North Dakota, South Dakota, and Minnesota. The data used in this research include patient emergency records collected between 2017 – 2019 from four large locations in the Midwest. Initially, the complete dataset has 478,212 records with 32 features. Those features explain all the events that happen during an ED visit including nurse checks, chief complaints, doctor diagnoses, etc. The following explains the criteria used to include/exclude the initial set of features:

- Since the goal of the proposed models is to predict patient admission status during a patient triage, only initial triage features are used to generate the prediction models. The factors that explain the events after a patient is triaged are excluded. This step reduces the number of features from 32 to 17.
- Patients whose disposition decision is other than admitted or discharged are excluded. Other disposition decisions include transferred patients, expired patients, and patients who refused treatment. This step reduces the records from 478212 to 453664.
- Timestamps are excluded except the arrival time. For example, the time between the arrival and discharge, the time between arrival, and the time a patient is first seen by a healthcare provider are excluded.
- The arrival time/date is converted into multiple features: month of the year (e.g., January, February), day of the week (e.g., Saturday, Sunday), and hour of the day (e.g., 1-24).
- Diagnosis information that happens after the triage is excluded.

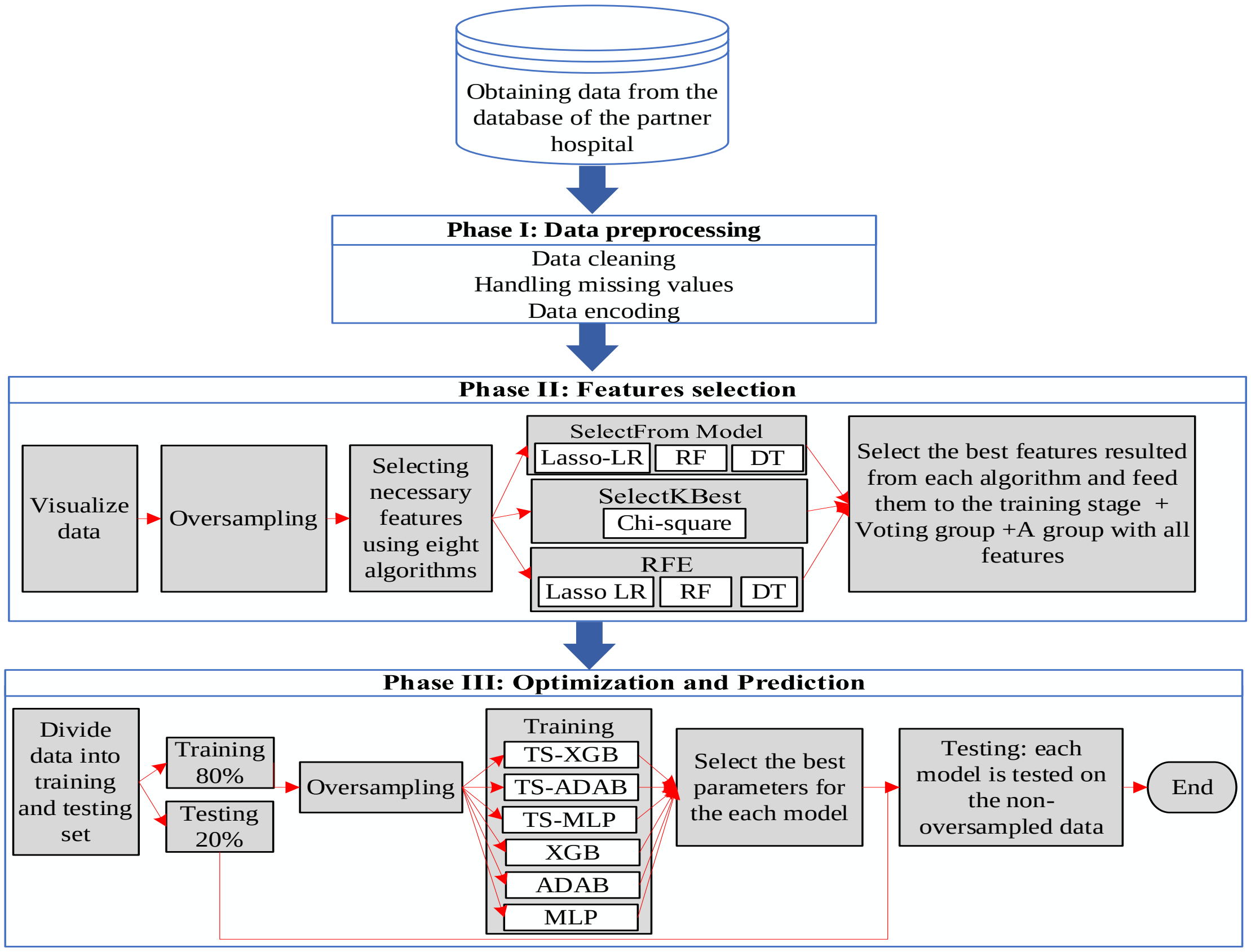

**Figure 1**: The research framework.

The final dataset has 17 features (See Table 1) and 453,664 patient records. The features can be categorized into two main groups: categorical and numerical features. The numerical features include BMI, patient age, diastolic blood pressure, temperature, pulse rate, respiratory rate, O2 saturation, and systolic blood pressure. The second group includes categorial features, which are patient sex, ethnicity, smoking status, and ED location ID. The day and the hour are treated as categorical features as well. The chief complaint feature is not visualized because there is a large number of categories that cannot be fitted in one chart.

**Table 1**: Features used for modeling.

| **Feature** | **Type** | **Feature** | **Type** |
|---|---|---|---|
| Patient Sex | Categorial, binary | BMI | Numerical |
| Ed Department Location ID | Categorial, integer | Age Years | Numerical |
| ED Arrival Time hour | Categorial, integer | Diastolic Blood Pressure | Numerical |
| Zip code | Categorial, integer | Temperature in Fahrenheit | Numerical |
| Patient Ethnicity | Categorial, integer | Respiratory Rate | Numerical |
| Patient Smoking Status | Categorial, integer | Pulse Rate | Numerical |
| month of year | Categorial, integer | Systolic Blood Pressure | Numerical |
| day of week | Categorial, integer | O2 Saturation | Numerical |
| Chief Complaint | Categorial, integer | | |

After cleaning and preprocessing the data, the sample size is 453,664 patient records and 17 features. However, the data includes lots of missing values. Table 2 shows the number and percentages of missing values for all the features that have missing values. The output feature (disposition decision) is plotted to understand the balance of the two classes (e.g., admitted and discharged). Figures 2 and 3 show the frequency charts of the two classes of the output feature before and after removing the missing values. Removing all the missing values results in a severe data imbalance problem. It negatively impacts the quality of the developed models. After examining the data, it is noticed that the missing values are not lab test results, thus using data imputation will not result in losing critical information.

Data imputation is performed using $k$-Nearest Neighbor (KNN) to reduce the effects of data imbalance and avoid losing valuable information due to removing all rows that have missing values. KNN imputer firstly calculates the Euclidean distances matrix for the observations. Then, it fills the missing observation with a value close to the neighbor observations, which is the average value of the neighbors. For example, if the number of neighbors is four, missing values will be the average of the four neighbors. After handling the missing data, categorical features such as gender, ethnicity, and smoking status are encoded using integer encoding, in which an integer number is given for each category in a feature. Then, two routes are taken to scale the data. If the algorithm used is tree-based (e.g., decision tree, XGB, ADAB), no normalization is applied to the data. However, with algorithms that have activation functions such as MLP and Lasso-LR, the input features are normalized before feature selection and model development. Since the number of records is very large (453,664), random sampling is conducted while building the models. A total of 5,000 random samples are withdrawn from the preprocessed dataset then the feature selection and model building are conducted.

**Table 2**: Percentage of missing values.

| Feature | No. of missing values | % of missing values |
|---|---|---|
| Respiratory Rate | 123203 | 27.2% |
| O2 Saturation | 122023 | 26.9% |
| Body Mass Index (BMI) | 116624 | 25.7% |
| Systolic Blood Pressure | 116624 | 25.7% |
| Diastolic Blood Pressure | 116624 | 25.7% |
| Pulse Rate | 116624 | 25.7% |
| Temperature in Fahrenheit | 116624 | 25.7% |
| Zip Code | 1065 | 0.0% |
| Remaining factures | 0 | 0.0% |

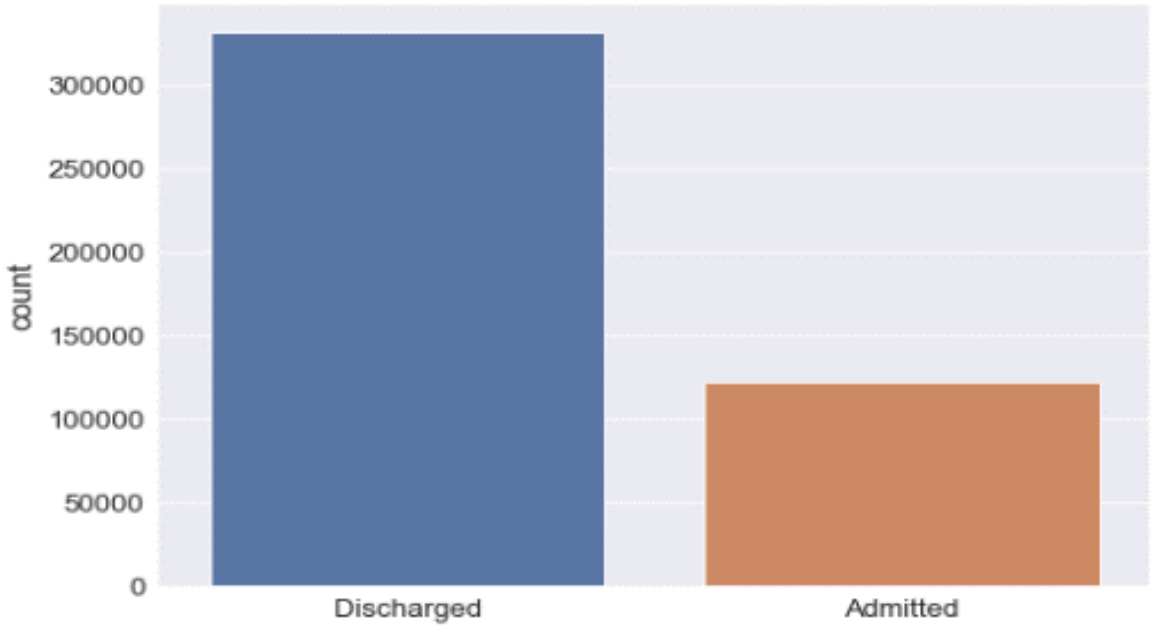

**Figure 2:** Disposition decision before removing missing values.

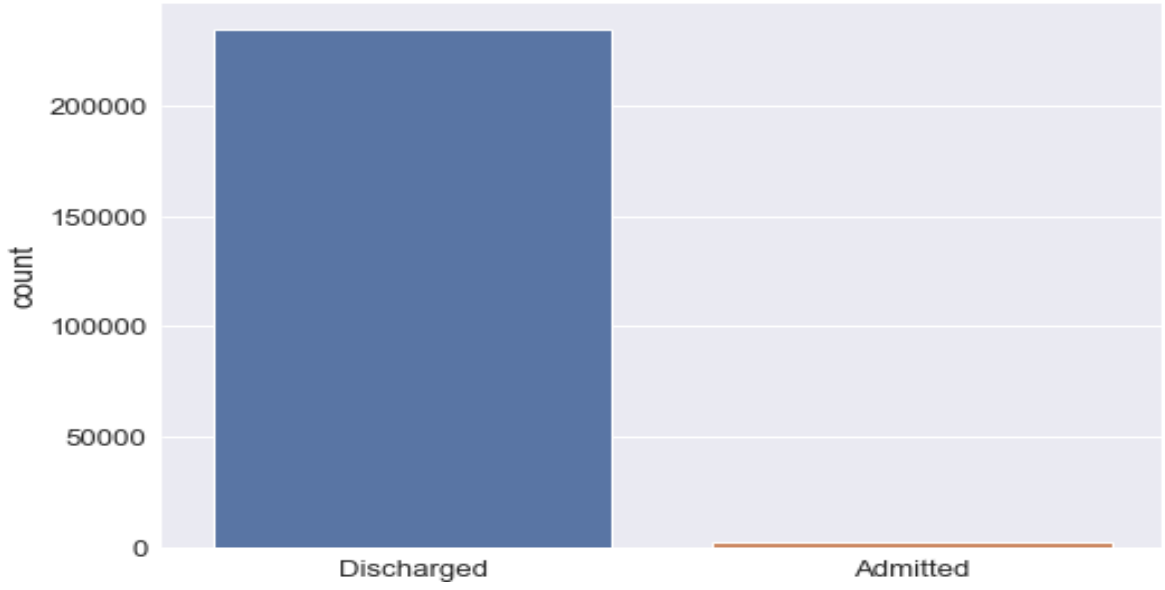

**Figure 3:** Disposition decision after removing missing values.

### 3.3 Feature Selection Methods

Before building the prediction models, a feature selection procedure is conducted. The purpose of the feature selection is to find a group in the dataset instead of using all the features to prevent overfitting and reduce computational complexity (Raza & Qamar, 2019). Four feature selection algorithms are used: Chi-sq, RF, DT, and LASSO-LR. The algorithms are executed using three Scikit-learn functions. The SKB function is utilized to implement Chi-sq feature selection method, while RFE and SFM functions are used to implement RF, DT, and LASSO-LR feature selection approaches. Therefore, seven data groups are obtained from the original data. A voting technique is also considered for feature selection. The voting group is created to include the features that are selected at least four times by the feature selection methods. If a feature is selected by only three or fewer methods, it is not added to the voting group. Another group includes all the 17 input features. Three feature selection algorithms are used: Lasso_LR, DT, and RF. In this work, Lasso-LR is implemented with SFM and RFE Scikit-learn functions. It is one of the regularization methods that eliminate unnecessary features. The shrinkage parameter (λ), in Lasso-LR is penalized as a model coefficient except for the intercept. As the unit $\lambda$ increases, the non-significant coefficient shrinks to a value equal to zero. Equation 1 is used to Lasso-LR, where $X$ represents variable inputs, $y$ represents the output, $\beta$ is the coefficient (Hastie et al., 2009). The data group that is obtained from the Lasso-LR method will be denoted by Lasso_SFM if it is obtained by the SFM function, and Lasso_RFE if it is obtained by the RFE function. Similar notation is used throughout the paper.

$$l_{\lambda}^{L}(\beta) = \sum_{i=1}^{N}\left[(y_i x_i \beta^T - \log\left(1 + e^{\beta^T x_i}\right)\right] - \lambda \sum_{j=1}^{p}\left|\beta_j\right| \tag{1}$$

Another feature selection method utilized in this paper is DT. It is also used as a base model in the ADAB prediction method. DT is a supervised, non-parametric, learning method. It has been used in classification (Kohavi & Quinlan, 2002), regression (Xu et al., 2005), and feature selection (Sugumaran et al., 2007). During the feature selection stage, it is implemented with RFE and SFM Scikit-learn functions. Using DT for feature selection relies on features' importance and the learning function applied (e.g., RFE or SFM). In RFE, all the features are included at the beginning and then a model removes the least important ones recursively. In SFM, a threshold is determined (e.g., mean of feature importance), then the features that are less than the threshold are removed and the others remain. Feature importance is calculated according to the Gini index, which measures the quality of a split when using a given feature. Equation 2 can be used to calculate the Gini index, given that D is a subset of a dataset, $n$ is the number of classes and $p_j$ is the portion of the samples labeled with class $j$ in the sample set $D$ (Han & Kamber, 2001). Throughout the paper, DT_SFM is used to denote the data group acquired from the SFM and the DT function, while DT_RFE is used to denote the group that is obtained from DT and RFE.

$$Gini(D) = 1 - \sum_{j=1}^{n} p_j^2 \tag{2}$$

This paper uses RF for feature selection. RF is similar to DT as a feature selection. The mode's feature importance (e.g., z-score) is the main criterion that is used to decide whether to include a feature or remove it. Suppose there are $B$ samples $(b = 1,2,\dots,B)$, $T$ trees, and $x_j$ $(j = 1,2,\dots N)$. To calculate the importance of $x_j$, Equations 3 and 4 are used to calculate the importance of variable $x_j$ $(j = 1,2, \dots N)$, which is $z_j$.

$$\bar{D}_j = \frac{1}{B} \sum_{b}^{B} \left( R_b^{oob} - R_{bj}^{oob} \right) \tag{3}$$

$$z_j = \frac{\bar{D}_j}{\frac{s_j}{\sqrt{B}}} \tag{4}$$

Where, $R_b^{oob}$ is the accuracy "out of bag" (OOB), $R_{bj}^{oob}$ is the permuted $x_j$ with the OOB data, $\bar{D}_j$ is the average decline in the accuracy of the classification $D_j$ for variable $x_j$, $s_j$ is the standard deviation, which is derived from the classification accuracy (Verikas et al., 2011).

The third feature selection approach used in the paper is Chi-sq and it is implemented using SKB Scikit-Learn function. In this approach, Chi-sq scores are calculated for each feature, and then the features that score the highest are designated while the rest are removed (e.g., SKB). Equation 5 is used to calculate the expected prevalence of samples in the $ith$ interval and $jth$ class, where $n$ reflects the number of samples and $r$ reflects the number of discrete intervals $r = (i = ,1,2, \dots r)$, $c$ reflects the number of classes $(c = 1, 2, \dots, c)$ and $n_{ij}$ is the actual frequency of samples in the $ith$ interval and $jth$ class. Discretization is applied for continuous features (Ma et al., 2017). Equation 6 is used to calculate the Chi-sq score. The data obtained from Chi-sq is denoted by Chi-sq_SKB.

$$\mu_{ij} = \frac{\sum_{j=1}^{c} n_{ij} \cdot \sum_{i=1}^{r} n_{ij}}{n} \tag{5}$$

$$\chi^2 = \sum_{i=1}^{r} \sum_{j=1}^{c} \frac{\left( n_{ij} - \mu_{ij} \right)^2}{\mu_{ij}} \tag{6}$$

SKB, RFE, and SFM Scikit-learn functions are used to implement RF, DT, and Lasso-LR for feature selection. The SKB functions use a statistical test to calculate scores for all features, then $K$ features that score the highest are designated and the rest are removed. In this study, the feature scores are calculated based on Chi-sq statistics. The RFE function begins with an initial group of factors which are then removed and included recursively based on their feature importance criterion. The SFM function selects features backward such that all features are included initially, and then insignificant ones are removed recursively until a model's performance is no longer improving.

## 3.4 The Proposed Machine Learning-Optimization Algorithms

In this work, a framework for optimizing XGB, ADAB, and MLP algorithms using TS is proposed. The goal of using TS is to boost the performance and increase the computational efficiency of the proposed three prediction algorithms that will be the core of an efficient decision tool to predict ED patient disposition decisions. The newly proposed algorithms are called: T-XGB, T-ADAB, and T-MLP where the "T" stands for Tabu search. This section presents the details of the proposed approach for optimizing predictive algorithms.

### *3.4.1 The Basic Idea*

Every machine learning algorithm has different hyperparameters and most of them have infinite possible values. For example, the value of the learning rate in a neural network model is between (0,1), but there is an infinite number of possible values to choose from. The problem of finding the optimal value of these hyperparameters can be considered an optimization problem. Suppose that a machine learning algorithm has $N$ hyperparameters, and the goal is to determine the optimal values of the given hyperparameters that maximize the objective values (e.g., accuracy). The optimization model can be expressed as follows:

$$Max\ f(\mathbf{x}) = f\ (x_1, x_i, \dots, x_n) \tag{7}$$

Subject to:

$$\psi_i^{lower} \le x_i \le \psi_i^{upper} \quad i = 1, 2, \dots N \tag{8}$$

Where, $f$(x) in Equation 7 is the objective function, $x_i$ is a machine learning parameter, and $\psi_i^{lower}$ and $\psi_i^{upper}$ are the lower and upper allowed values for the parameter $x_i$, respectively. Equation 7 represents the objective function, while Equation 8 is the constraint set. The objective is to determine the optimal combination parameters **x** to maximize the prediction while satisfying the boundary constraints of each parameter. The values of $\mathbf{x}$ can be an integer, float, or binary. For example, in XGB, the learning rate parameter must be between [0, 1], while the maximum delta step is an integer and between [0, ∞). Such constraints cannot be violated while selecting an algorithm parameter. In this work, the TS algorithm is used to maximize the hyperparameters of three algorithms: T-XGB, T-ADAB, and T-MLP.

### *3.4.2 Tabu Search for Optimizing Machine Learning Models*

TS is a neighborhood search algorithm that was introduced by Glover in 1986 (Glover, 1986). It has been broadly used to calculate various combinatorial maximization problems due to its efficiency and simplicity (Gendreau, Michel, Potvin, 2019). The algorithmic structure of TS includes four main components: 1) Tabu list (TL), which is used to keep track of recently visited solutions to avoid cycling; 2) aspiration criteria, which allows a solution to be explored even if it violates the tabu constraint when the solution results in a better score than the current solution; 3) intensification, which allows the algorithm to go back to the best solution if the search space is not promising; and 4) diversification strategy, which guides the search to an unvisited solution. These strategies prevent the algorithm from falling into the local-maxima trap and help explore solutions in different search spaces to obtain better solutions. Another important component of the TS algorithm is the type of memories utilized in the search, which include both short and long-term memories. While short-term memory prevents the algorithm from applying moves that have already been visited, long-term memory keeps track of good solutions.

The steps of optimizing T-ADAB, T-XGB, or T-MLP using TS are shown in Figure 4. In step 1, a preliminary result $S^*$ is generated and $S^*$ is set to be the current solution $S_{curr}$. The preliminary result ($S^*$) is generated based on a Uniform distribution, where the minimum limit is the lowest possible value for a parameter and the maximum value is chosen to be a small value to avoid local maxima trap. For example, the number of estimators for a T-XGB is generated between (1, 5), while for learning rate is generated between (0.001, 0.1). In step 2, the initial solution $S^*$ becomes the current solution. An example, of a solution representation for a T-XGB model, is shown in Figure 5. It is represented by a one-dimension array where each element in the array represents the value for a parameter. For example, the first element in the array represents the number of estimators, while the second element represents the max depth of the T-XGB tree. Then, a model (e.g., T-XGB, T-ADAB, or T-MLP) is trained and tested based on $S^*$. The AUC of testing a model is the main criterion for evaluation. The values of the parameters of any of the three algorithms (e.g., T-XGB, T-ADAB, or T-MLP) are generated from a uniform distribution for the initial solution $S^*$. The upper and lower values of the uniform distribution are set to be low to avoid overfitting from the beginning of the search. For example, the initial learning rate is generated from a uniform distribution with upper and lower bounds of 0.01 and 0.1, respectively. Then, the normal distribution is used to generate neighborhood solutions from $S^*$, which is explained in Step 3.

In step 3, neighborhood solutions, $N(S)$, are generated. The moving operator to search a neighborhood for each parameter value is set by increasing or decreasing the value of a parameter based on a normal distribution. A Normal distribution's mean and standard deviation are determined in two ways. If the value of a parameter can be greater than 1 (e.g., number of estimators in the T-XGB), the mean and standard deviation values are equal to 0, 2, respectively (Figure 6). If the value of a parameter is between 0 and 1 (e.g., learning rate), the mean and standard deviation are equal to 0, and 0.1, respectively (Figure 7). If the value of the random number generated by the Normal distribution is positive, the current value of a parameter is increased, otherwise, it is decreased. For example, suppose that the current value of a learning rate is 0.05 and the randomly generated number from a normal distribution is $0.008$, then the next value of the learning rate will be $0.05 + 0.008 = 0.058$. In another iteration, the learning rate may decrease depending on the number generated from the Normal distribution. The same applies to the other parameters. This applies to all three algorithms: T-XGB, T-ADAB, and T-MLP. In steps 4 and 5, each solution in the neighborhood $N$ is checked if it violates an algorithm constraint, knowing that all the parameters have boundary constraints. For example, in the T-XGB, the maximum number of estimators is set to be 50. Therefore, while generating neighborhood N, the value of the estimator either decreases or increases but cannot be greater than 50. If a parameter violates its constraints, it is fixed by bringing it to the range of its value.

In steps 6 and 7, a model (e.g., T-XGB, T-ADAB, or T-MLP) is trained and tested for each solution $S$ in $N$. Then, the solution with maximum AUC is selected ($S_{best}$) in step 8. In step 9, $S_{best}$ is checked if it is not in the TL. Then, if that is true, the solution $S_{best}$ is added to the TL (step 10) and it becomes the $S_{curr}$ (step 11). In step 12, the good solutions are tracked and stored in the long-term memory. If $S_{best}$ is already in the TL, the algorithm keeps track of good solutions (step 12), select the next best candidate $S_{best}$ (Step 13) and repeat steps 9-13. Each time a solution is added to the TL, the length of the TL is checked if it is greater than the maximum allowed length and if that happens, the oldest solution in the TL is deleted (steps 14, 15). In step 16, the diversification strategy is applied by generating a completely new random solution based on a small probability and used as $S_{curr}$. This process continues until stopping criteria are met (Step 17). This study's stopping criterion is the maximum total number of iterations, which is 300.

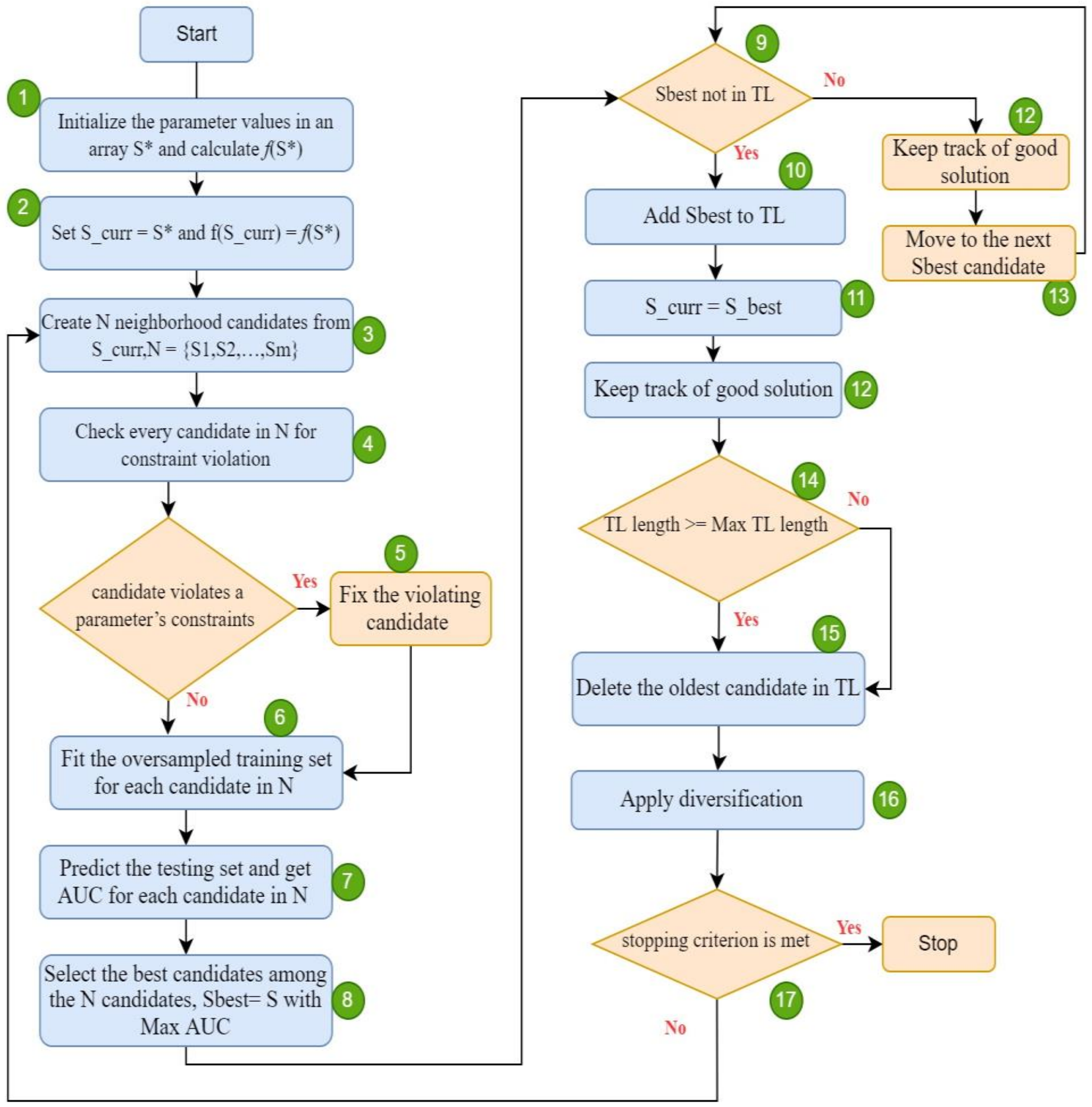

**Figure 4:** Flow chart for optimizing T-XGB, T-ADAB, and T-MLP.

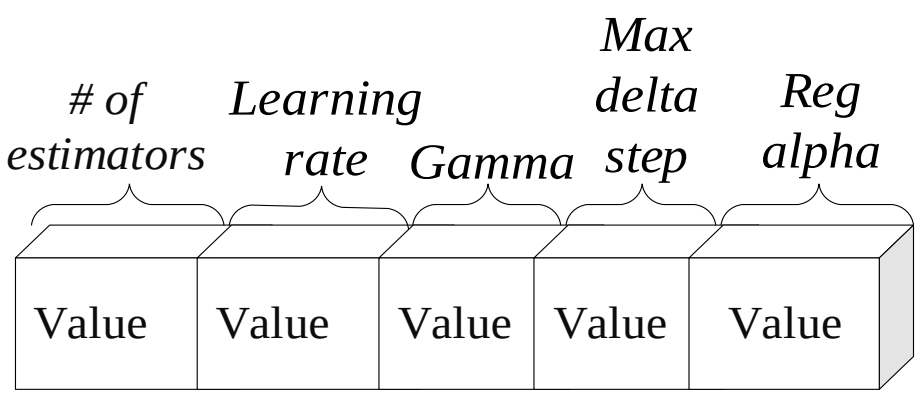

**Figure 5**: representation of the parameters of an XGB model.

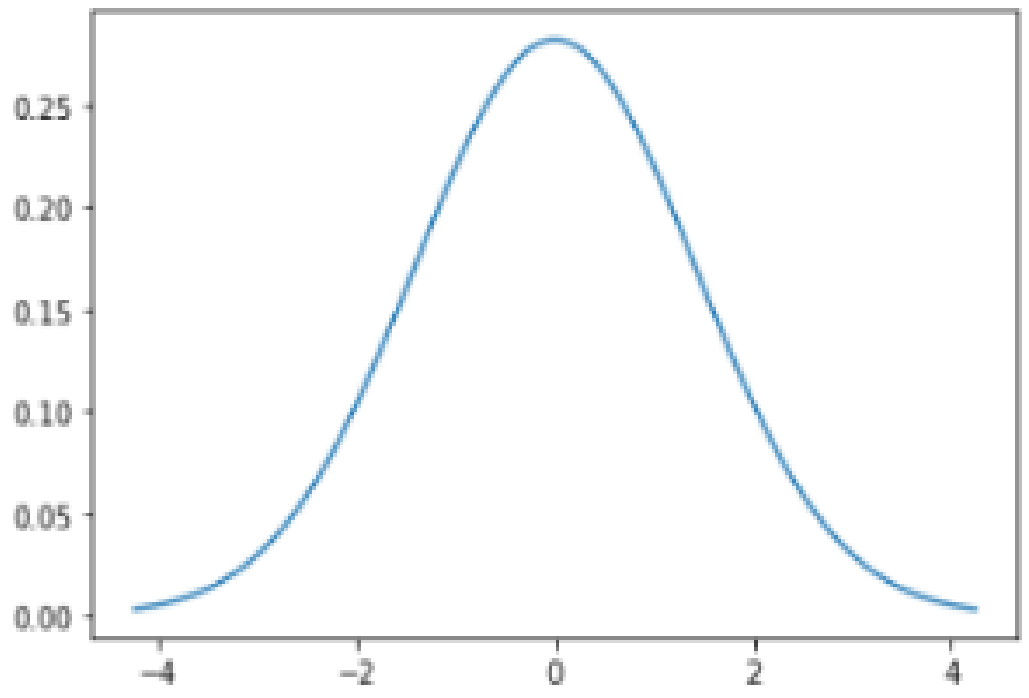

**Figure 6:** Neighborhood search distribution for parameters with values greater than 1.

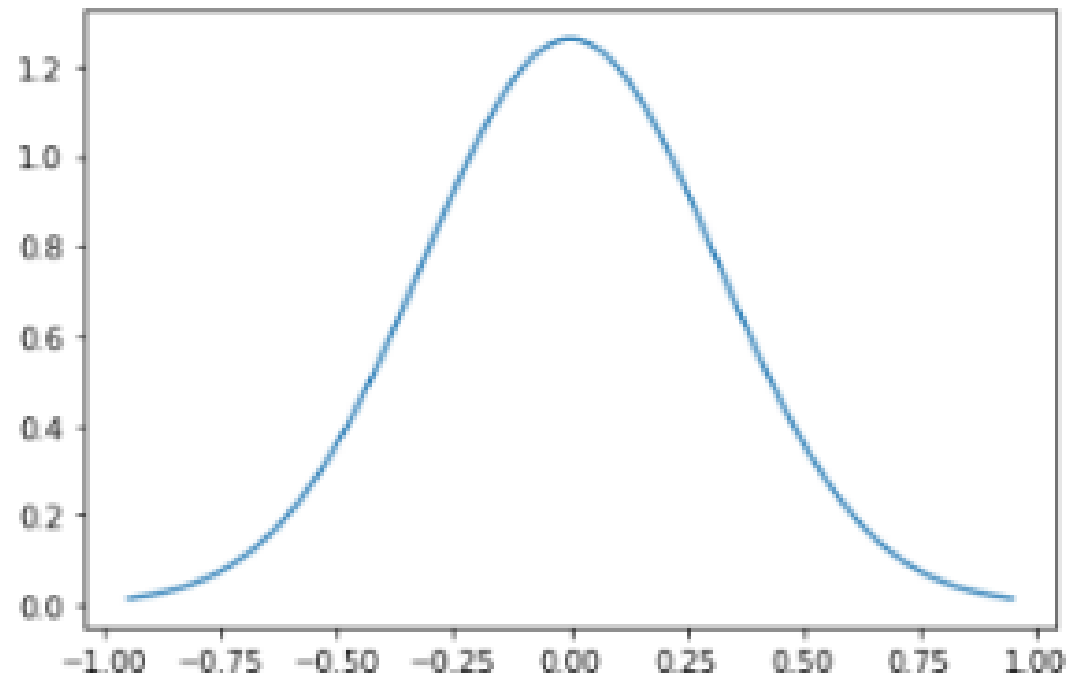

**Figure 7:** Neighborhood search distribution for parameters values between (0,1).

### *3.4.3 Tabu Search-AdaBoost*

Adaptive Boosting or AdaBoost (ADAB) is an ensemble machine learning algorithm and one of the boosting algorithms. It combines weak learners to create a collectively strong learner. The original data is first trained using a base learner (e.g., decision tree). Then, the base learner is trained on weighted data where misclassified from the previous stage are given higher weights in the current stage. Increasing the weight of misclassified instances helps in increasing the instances of correct classifications in the next iteration. The process is replicated *n* times until the desired performance is reached (Kumar & Jain, 2020). In this work, TS is used to optimize the ADAB (T-ADAB) hyperparameters and their base learner. Two T-ADAB hyperparameters and three parameters for the base learner are considered. The parameters are the T-ADAB number of estimators, T-ADAB learning rate, maximum depth of the base learner, the minimum samples split of the base learner, and the minimum sample leaf of the base learner, knowing that the base learner is DT.

### *3.4.4 Tabu Search-Extreme Gradient Boosting (XGB)*

Extreme gradient boosting (XGB) has been used in feature selection (Chen et al., 2020), classification (D. Yu et al., 2020), and regression (Chen et al., 2015). In XGB, a single strong learning model is built based on small weak learners. In general, a model that is based on one tree is a weak learner, but combining multiple ones generates a strong learner. In XGB, trees are created recursively, where misclassified instances from a previous step are given higher weight in a current instance. XGB is similar to ADAB except that in ADAB new weak learners are added after increasing the weights of misclassified instances while in XGB a model is trained on residual errors made by a previous learner. In other words, weak learners in XGB are generated based on optimizing the lost function of decision trees. In this paper, TS and XGB (T-XGB) are integrated to create a robust prediction algorithm with optimal hyperparameters. T-XGB has 21 parameters and most of them have infinite possible values, but in this work, six parameters are considered for optimization. The parameters are the number of estimators, maximum depth, learning rate, gamma, maximum delta step, and the number of parallel trees. More information about XGB hyperparameters can be found in (*XGBoost Parameters — Xgboost 1.4.0-SNAPSHOT Documentation*).

### *3.4.5 Tabu Search-Artificial Neural Network*

The ANN is a supervised machine learning algorithm that models the functionality of the human brain. It is constructed from artificial neurons that are constructed in a series of layers. There are multiple variations of ANN algorithms. This paper uses the multilayer perceptron (MLP), which is constructed from input, hidden, and output layers. The input layer represents the input features, while the output layer represents the output variable. The number of hidden layers varies in terms of the number of layers and the number of nodes in each layer (Niel & Bastard, 2019). A large number of hidden layers increases the computation complexity and may lead to overfitting, while a small number of hidden layers may lead to underfitting. To handle nonlinearity in data, an activation function is used to transform the data from one layer into the next one. Since our response variable is binary and the input variables have binary and non-binary features, this paper uses the sigmoid activation function, which guarantees that the output variable (e.g., a class probability) is between 0 and 1 (Jamel & Khammas, 2012). Further, TS is used to optimize its parameters and to obtain the best performance of MLP. Three hidden layers are considered and then the number of nodes in the three layers, learning rate, momentum, and alpha are optimized. More information about MLP parameters can be found in Scikit-learn documentation (*Sklearn.Neural_network.MLPClassifier — Scikit-Learn 0.24.0 Documentation*).

## 3.5 Performance Measures

Five performance measures are used to assess the proposed prediction algorithms. The performance measures included are accuracy (Equation 7), sensitivity (Equation 8), specificity (Equation 9), F1 score (Equation 10) (Han & Kamber, 2001), and the area under the curve (AUC). In the dataset, the class of an admitted patient is labeled as zero (negative) and a discharged patient is labeled as one (positive). Therefore, a true positive (TP) instance occurs when a model correctly predicts class 1 (discharged patient), while a false negative (FN) occurs when the model misclassifies predicts class 1 (discharged patient). True negative (TN) occurs when the model truly predicts class 0 (admitted patient), while false positive (FP) instance occurs when the model does not classify class 0 (admitted patient) correctly. Since the two classes are imbalanced, the AUC is used as the main performance measure to determine which model is the best. AUC is the area the sensitivity and (1 – specificity).

$$Accuracy = \frac{\mathrm{TP} + \mathrm{TN}}{TP + FN + FP + FN} \tag{7}$$

$$Sensitivity = \frac{\mathrm{TP}}{TP + FN} \tag{8}$$

$$specificity = \frac{\mathrm{TN}}{TN + FP} \tag{9}$$

$$Precision = \frac{TP}{TP + FP} \tag{10}$$

$$F1 = 2 * \frac{Precision * sensitivity}{Precision + sensitivity} \tag{11}$$

# 4. EXPERIMENTAL RESULTS

This section provides the findings of the proposed prediction models, in addition to the optimization model.

## 4.1 Feature Selection Results

A combination of seven feature selection algorithms is used to determine the best features among the 17 features, which results in seven different subsets. Two additional groups are considered; one is based on voting and the other included all features. Table 3 shows the features selected by each selection method. The last column in Table 3 represents the selection frequency of each feature. For example, the O2 saturation feature was designated as significant by all the feature selection methods. Therefore, the total selection is six. Moreover, patient age is selected by six feature selection methods. Chief Complaint is selected by three methods, so it is not considered in the voting. Table 3 gives a complete understanding of the features affecting patient admission. For example, the features that were removed by an algorithm are unimportant with respect to the patient admission status. On the other hand, the frequently selected features reflect greater importance with respect to the ED patient admission status.

**Table 3**: Feature selection results.

| No. | Feature | Lasso_ SFM | DT_ SFM | RF_ SFM | Chi_ SKB | DT_ RFE | RF_ RFE | Lasso_ RFE | Voting | Total |
|---|---|---|---|---|---|---|---|---|---|---|
| 1 | O2 Saturation | √ | √ | √ | | √ | √ | √ | √ | 6 |
| 2 | Age Years | √ | | √ | √ | √ | √ | √ | √ | 6 |
| 3 | Systolic Blood Pressure | √ | √ | | √ | √ | √ | √ | √ | 6 |
| 4 | BMI | √ | √ | | √ | √ | √ | √ | √ | 6 |
| 5 | Respiratory Rate | | √ | √ | √ | √ | √ | | √ | 5 |
| 6 | Pulse Rate | √ | √ | √ | | √ | | √ | √ | 5 |
| 7 | Zip code | √ | | √ | | √ | √ | √ | √ | 5 |
| 8 | Diastolic Blood Pressure | √ | | | √ | | √ | √ | √ | 4 |
| 9 | Patient Sex | √ | | √ | √ | | | √ | √ | 4 |
| 10 | Chief Complaint | | | | √ | √ | √ | | | 3 |
| 11 | Ed Department Location ID | | | | √ | √ | √ | | | 3 |
| 12 | Patient Ethnicity | √ | √ | | | | | √ | | 3 |
| 13 | Temperature in Fahrenheit | √ | | | | | | √ | | 2 |
| 14 | ED Arrival Time hour | | | √ | | | √ | | | 2 |
| 15 | Patient Smoking Status | | √ | | √ | | | | | 0 |
| 16 | Month of year | | | | √ | √ | | | | 0 |
| 17 | Day of week | | | | | | | | | 0 |

## 4.2 Optimization Settings

To optimize a machine learning algorithm using metaheuristics, the bounds of each parameter should be determined. Table 4 presents the possible range of each parameter for T-XGB, T-ADAB, and T-MLP. Some of these parameters have no upper limit, thus an upper limit of such parameters is specified to reduce the search space for each parameter, avoid overfitting, and improve computational efficiency. Table 5 shows TS parameters that are set for optimizing the three algorithms (e.g., XGB, ADAB, and MLP). The initial solutions for the three algorithms are generated based on a Uniform distribution while neighbors for each solution are obtained from a Normal distribution. In each iteration, a random number is calculated and if it is less than the probability of diversification, a new solution array is generated.

**Table 4:** Parameter ranges for T-XGB, T-ADAB, and T-MLP algorithms.

| Algorithm | Parameter | Possible range | Experimental setting | Type |
|---|---|---|---|---|
| T-XGB | Number estimators | [1,∞] | [1,50] | Integer |
| | Max depth | [0,∞] | [0,50] | Integer |
| | Learning rate | [0,1] | [0,1] | Float |
| | Gamma | [0,∞] | [0,50] | Float |
| | Max delta step | [0,∞] | [0,50] | Integer |
| | Number of parallel trees | [0,∞] | [0,50] | Integer |
| T-ADAB | Number estimators ADAB | [1,∞] | [1,50] | Integer |
| | Learning rate ADAB | [0,1] | [0,1] | Float |
| | Max depth base learner (DT) | [1,∞] | [1,50] | Integer |
| | Min samples split base learner (DT) | [1, sample size] | [1, 50] | Float |
| | Min sample leaf base learner (DT) | [1, sample size] | [1, 50] | Integer |
| T-MLP | Hidden layer sizes | [1,∞] | [1,30] | Integer |
| | Learning rate | (0, 1] | (0, 1] | Float |
| | Momentum | (0, 1] | (0, 1] | Float |
| | Alpha | (0,1] | [1, 1] | Float |

**Table 5**: TS parameters.

| **TS parameter** | **Value** |
|---|---|
| Number of iterations | 300 |
| Probability of Diversification | 0.002 |
| Tabu list length | 20 |
| Initial solution generation | Uniform distribution |
| Neighborhood search | Normal distribution |

## 4.3 Optimization Results

After setting the parameters of the three algorithms (T-ADAB, T-XGB, and T-MLP), TS is used to optimize each model resulted from the feature selection step and for all features. In other words, TS optimizes nine T-XGB models, nine T-ADAB models, and nine T-MLP models. TS is run for 300 iterations for all the models. In each iteration, a model is trained, tested, and the AUC is obtained as it is considered as the main performance measure. Figures 8 – 10 show the convergence for the T-XGB, T-ADAB, and T-MLP, respectively. Most of the models converge after iteration number 250 for T-ADAB, T-XGB, and T-MLP. The T-ADAB resulted in the best AUC, followed by T-XGB, and then T-MLP. The AUCs of all the T-ADAB models range between 87% and 95.4%, while the AUCs of the T-XGB models are between 88.8% and 94.8%. For the T-MLP, the AUCs of the optimal models are between 80.8% and 87.9%. The variations of the T-ADAB models for different data groups are larger than the variations of the T-XGB models and T-MLP models. More specifically, three out of the nine T-ADAB models have AUCs of less than 90%. The T-XGB models are above 90% except for one model which is derived from the Chi_SKB data group. With regard to the T-MLP algorithm, all its models are below 90%, except the model that resulted from the RF_RFE data group. T-ADAB results in the best model with an AUC of 95.4% and it is derived from the data group of RF_RFE.

Table 6 presents the optimal parameters for T-XGB. The best T-XGB model results from the DT_SFM data group. Looking at the feature selection algorithm regardless of the Scikit-learn function, the best AUCs for T-XGB models results from the data group obtained from DT, followed RF, X_all, then voting. With regard to the optimal parameters for the T-XGB best model, it can be noticed that the number of estimators is the highest for the best model. Conversely, the optimal depth and number of parallel trees for the best model are not the highest compared with the other T-XGB models. The learning rate for the best model is not the highest for the optimal model as well, while gamma is the second-highest value among T-XGB models. Table 7 presents the optimal hyperparameters for all T-ADAB models obtained after running the TS algorithm. The table includes the optimal parameter of T-ADAB and its base model (DT). The first two parameters belong to T-ADAB and the rest belong to the base learner. The model with the highest performance among T-ADAB is the model that results from the RF_RFE data group, which is the best model among all the 27 developed models. The optimal parameters for the best T-ADAB model have the highest values compared with other models. For example, the number of estimators and the learning rate for the optimal model is the highest compared with the other ADAB models. For T-MLP optimal models, the optimal parameters are shown in Table 8. The best T-MLP model is based on the data group resulted from Lasso_SFM.

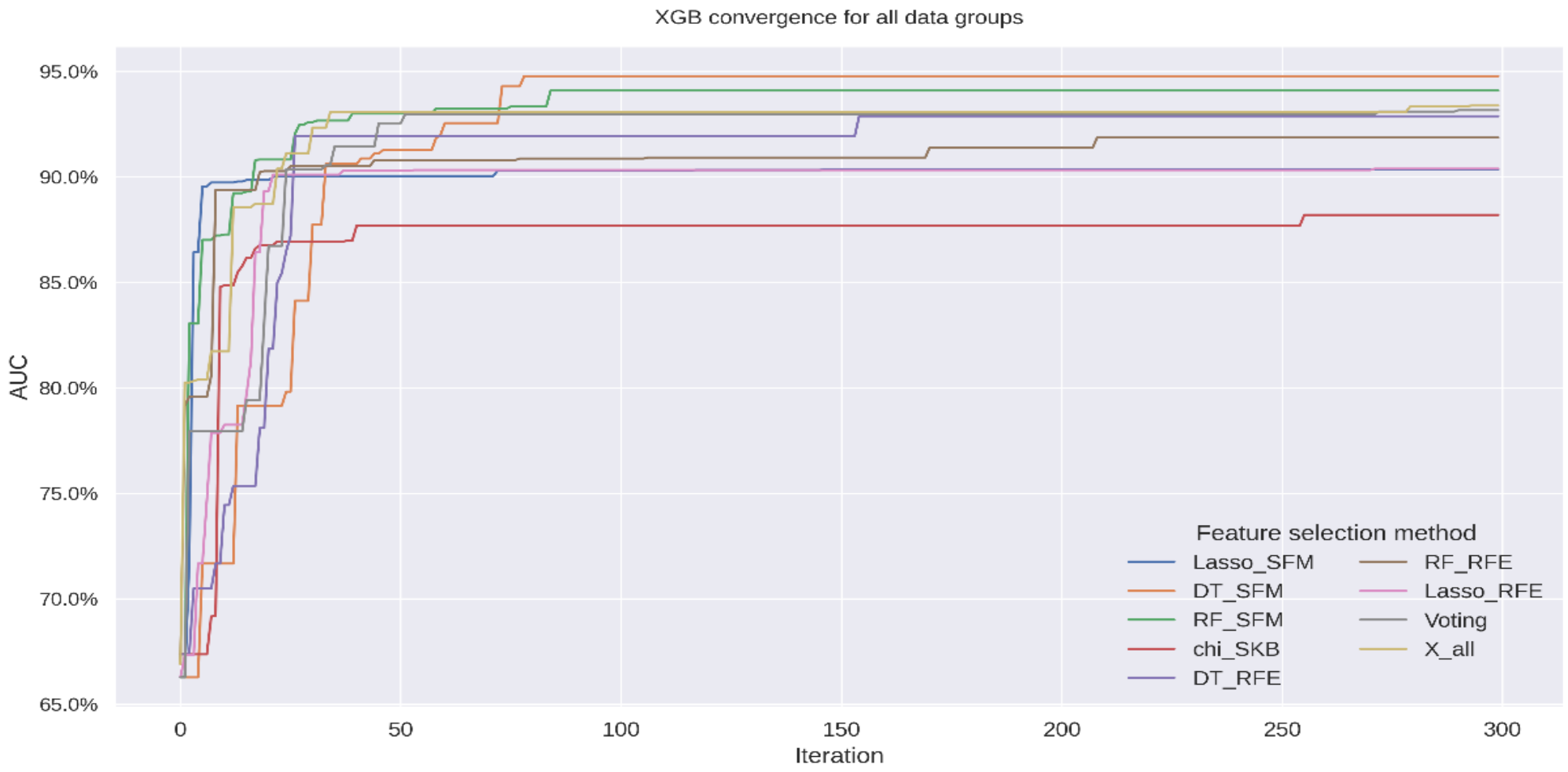

**Figure 8:** Convergence of T-XGB models for all data groups.

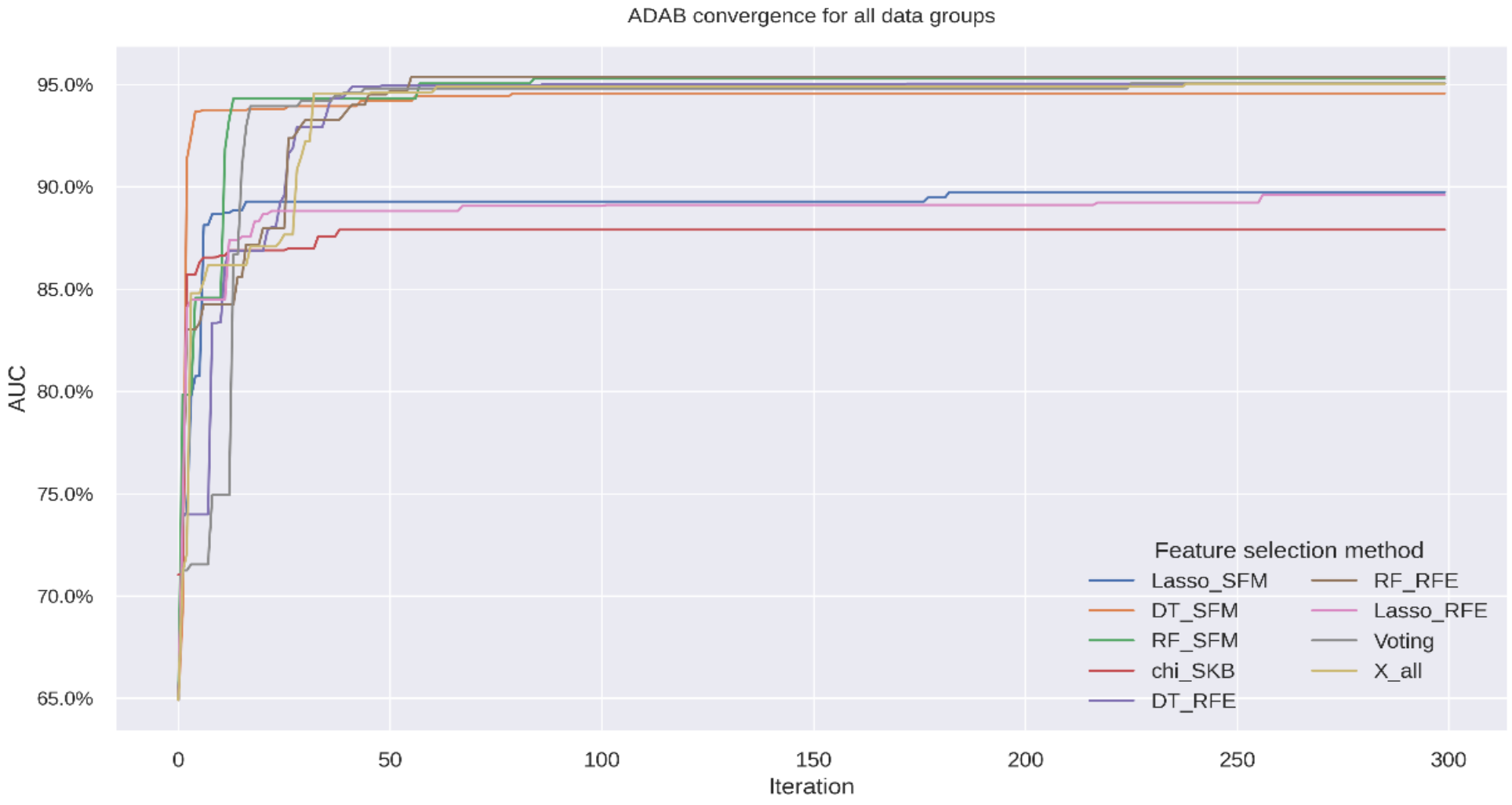

**Figure 9:** Convergence of T-ADAB models for all data groups.

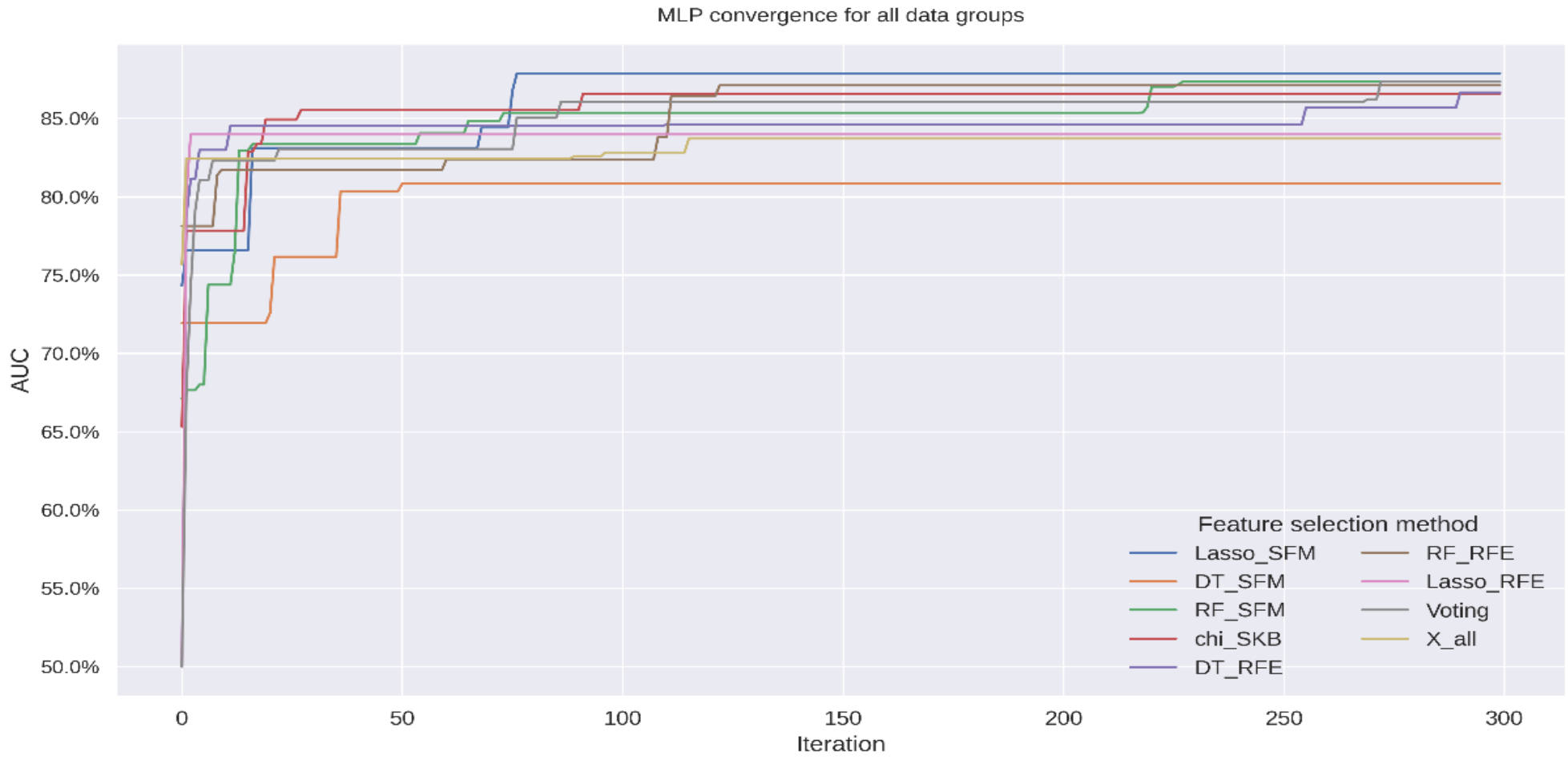

**Figure 10:** Convergence of T-MLP models for all data groups.

**Table 6:** Optimal parameters T-XGB and corresponding optimal AUC.

| T-XGB parameter | Model | | | | | | | | |
|---|---|---|---|---|---|---|---|---|---|
| | Lasso_ SFM | DT_ SFM | RF_ SFM | Chi_ SKB | DT_ RFE | RF_ RFE | Lasso_ RFE | Voting | X_all |
| # of estimator | 3 | 14 | 9 | 11 | 14 | 10 | 2 | 10 | 10 |
| Max depth | 12 | 23 | 5 | 13 | 7 | 8 | 15 | 4 | 5 |
| Max delta | 6 | 3 | 3 | 1 | 1 | 4 | 8 | 2 | 6 |
| # parallel tree | 8 | 1 | 7 | 1 | 7 | 2 | 1 | 1 | 1 |
| Learning rate | 0.023 | 0.075 | 0.184 | 0.114 | 0.059 | 0.038 | 0.016 | 0.116 | 0.167 |
| Gama | 1.481 | 1.621 | 1.792 | 0.578 | 1.883 | 1.078 | 0.734 | 0.601 | 0.223 |
| Optimal AUC | 90.4% | 94.8% | 94.1% | 88.2% | 92.9% | 91.9% | 90.4% | 93.2% | 93.4% |

**Table 7:** Optimal parameters T-ADAB and corresponding optimal AUC.

| T-ADAB and base learner parameter | Model | | | | | | | | |
|---|---|---|---|---|---|---|---|---|---|
| | Lasso_ SFM | DT_ SFM | RF_ SFM | Chi_ SKB | DT_ RFE | RF_ RFE | Lasso_ RFE | Voting | X_all |
| # estimators T-ADAB | 10 | 2 | 2 | 11 | 8 | 11 | 2 | 7 | 8 |
| Learning rate T-ADAB | 0.102 | 0.208 | 0.010 | 0.010 | 0.010 | 0.276 | 0.010 | 0.032 | 0.221 |
| Max depth DT | 3 | 12 | 9 | 13 | 8 | 15 | 10 | 7 | 19 |
| Min samples split DT | 9 | 15 | 11 | 14 | 14 | 15 | 6 | 19 | 18 |
| Min samples leaf DT | 12 | 15 | 7 | 19 | 20 | 11 | 13 | 12 | 12 |
| Optimal AUC | 89.7% | 94.6% | 95.3% | 87.9% | 95.0% | 95.4% | 89.6% | 95.1% | 95.0% |

**Table 8:** Optimal parameters T-MLP and corresponding optimal AUC.

| | Model | | | | | | | | |
|---|---|---|---|---|---|---|---|---|---|
| T-MLP parameter | Lasso_ SFM | DT_ SFM | RF_ SFM | Chi_ SKB | DT_ RFE | RF_ RFE | Lasso_ RFE | Voting | X_all |
| # of nodes - hidden layer#1 | 1 | 3 | 6 | 1 | 4 | 3 | 2 | 1 | 6 |
| # of nodes - hidden layer#2 | 8 | 1 | 4 | 8 | 2 | 7 | 2 | 7 | 9 |
| # of nodes - hidden layer#3 | 3 | 4 | 6 | 1 | 1 | 7 | 1 | 3 | 1 |
| Learning rate | 0.000 | 0.019 | 0.014 | 0.034 | 0.042 | 0.010 | 0.038 | 0.012 | 0.046 |
| Alpha | 0.071 | 0.043 | 0.063 | 0.028 | 0.014 | 0.072 | 0.043 | 0.128 | 0.081 |
| Momentum | 0.068 | 0.006 | 0.096 | 0.056 | 0.044 | 0.069 | 0.063 | 0.115 | 0.050 |
| Optimal AUC | 87.9% | 80.8% | 87.4% | 86.6% | 86.7% | 87.1% | 84.0% | 87.4% | 83.7% |

## 4.4 Prediction Results

This section presents the prediction results of the proposed models. The AUCs of the testing stage for the optimal models are shown in Figure 11. The sensitivity, specificity, f1-score, and accuracy are shown in Figures (12– 15), respectively. Every line chart represents the performance of an algorithm. The x-axes represent the feature selection method, while the y-axes represent the models' performance. Most of the optimized T-ADAB and T-XGB models resulted in AUCs over 90% (Figure 11). More than 17 models out of the 54 models have AUCs larger than 90%. The T-ADAB model that is built from the RF_RFE data group resulted in the highest AUCs, which is 95.4%. The second highest AUCs also resulted from T-ADAB but with a different feature selection method, which is DT_SFM, with an AUC of 95.3%. T-XGB also resulted in a model with a close AUC to the best model. It is based on the DT_SFM and RF_SFM data groups with AUCs of 94.8% and 94.1%, respectively. Since AUC is considered the main performance measure in this work, the model that is derived from that data group obtained from RF_REF and optimized T-ADAB is considered to be the best model (AUC of 95.4%). The best model resulted in the highest sensitivity (99.3%), specificity (91.4%), F1 (95.2%), and accuracy (97.2%). Therefore, the best model can predict the admission status (e.g., admitted vs. discharged) with high performance. Our final and best model (RF_RFE_T-ADAB) outperformed the models cited in the previous work. For example, Fernandes et al. (2020) presented a review paper about machine learning applications in improving ED operations. About 10 studies were included about machine learning usage in predicting admission decisions. The accuracy of those models ranged between 88% - 92%. A recent study published by De Hond et al. (2021) and they developed machine learning models to predict admission disposition at different stages including triage, after 30 min, and after 1 hour. The model that was based on triage information resulted in an AUC of 86%.

With regard to the prediction results of the optimized algorithms regardless of feature selection methods, it can be seen that optimized T-ADAB resulted in the best testing results, followed by T-XGB, and then MLP. The few T-ADAB hyperparameters could be the reason that it performs better than T-XGB. T-ADAB is also robust to overfitting in a low noise database (Rätsch et al., 2001). Although T-ADAB performed better than T-XGB, the differences among AUCs for all optimal T-ADAB and T-XGB models are not large. They are all within the range of 88% – 95% and most of them are larger than 90%. The performances of the traditional MLP models are the worse among all the optimal models of the three algorithms. Comparing the optimized models (e.g., T-XGB, T-ADAB, and T-MLP) with the traditional models, it can be seen in Figures 11 – 15 that the optimized models outperformed all the traditional models in all performance measures.

Figure 16 shows the feature significance for the best-performing model, which is derived from the F-score of the features. The features are ranked from the most important to the least important with respect to their effects on the admission status of ED patients. The most important feature is O2 saturation, while the least important is the location of ED feature, knowing that the data of this study is collected from three locations of the partner hospital.

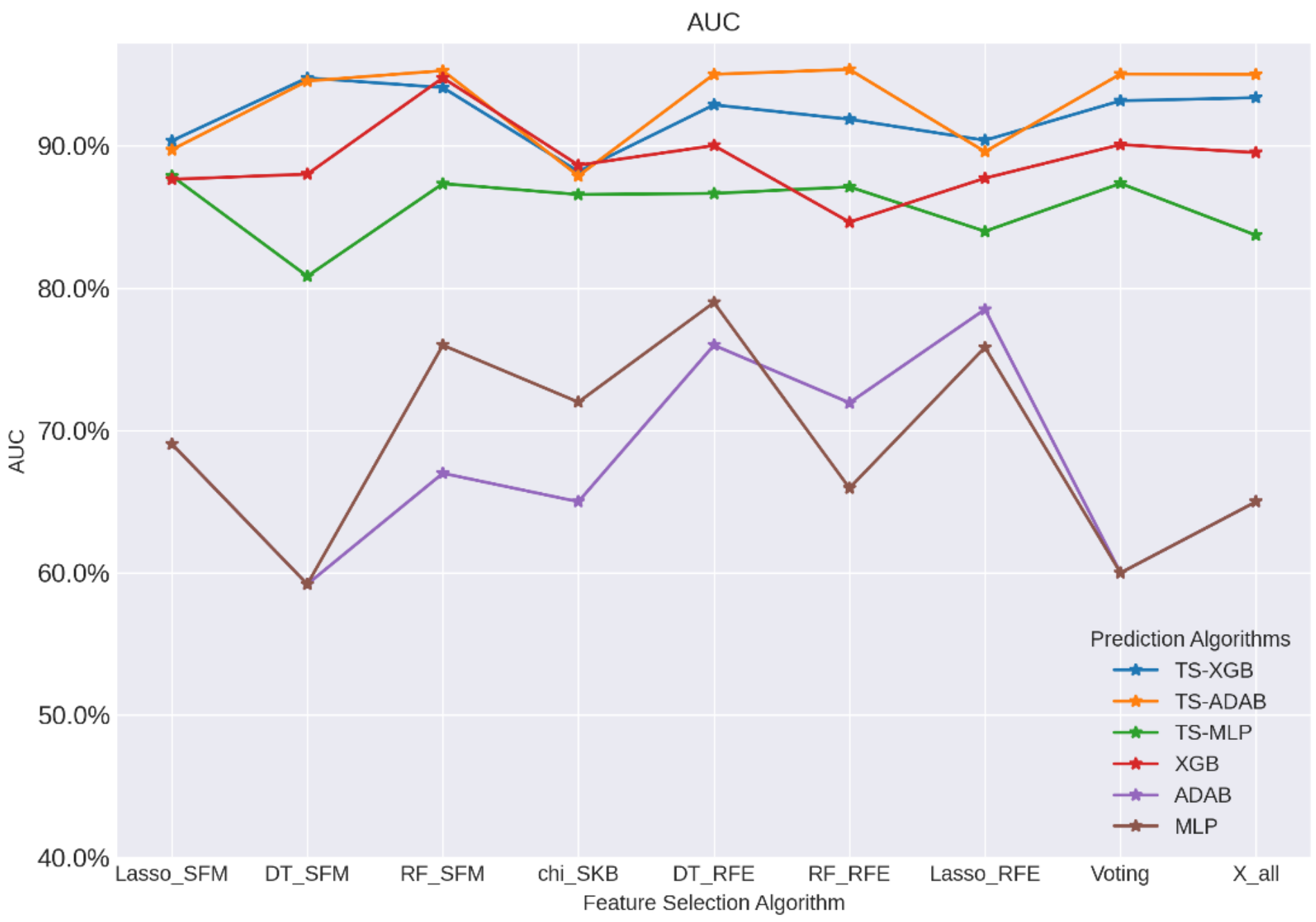

**Figure 11**: AUC for all models.

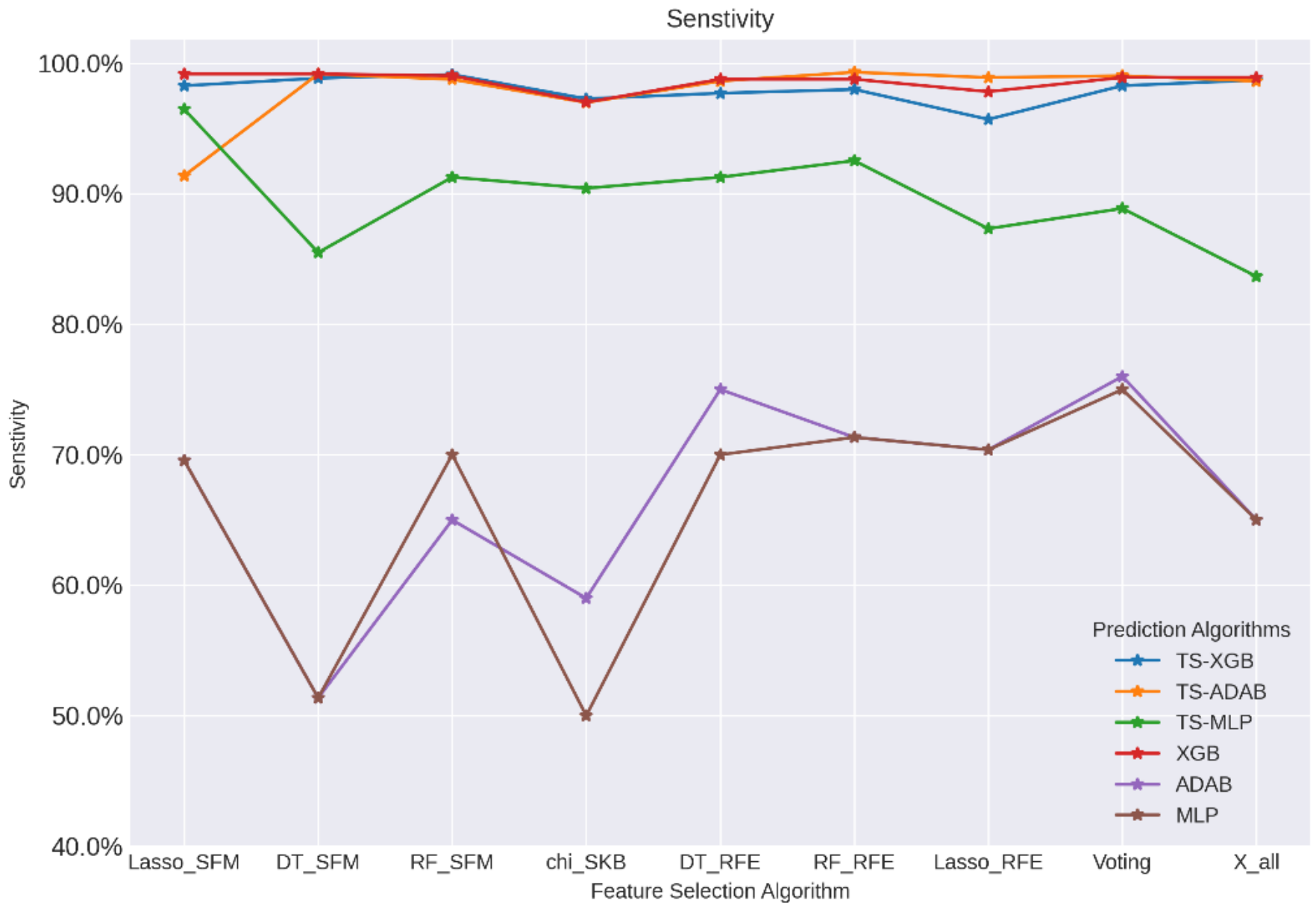

**Figure 12:** Sensitivity for all models.

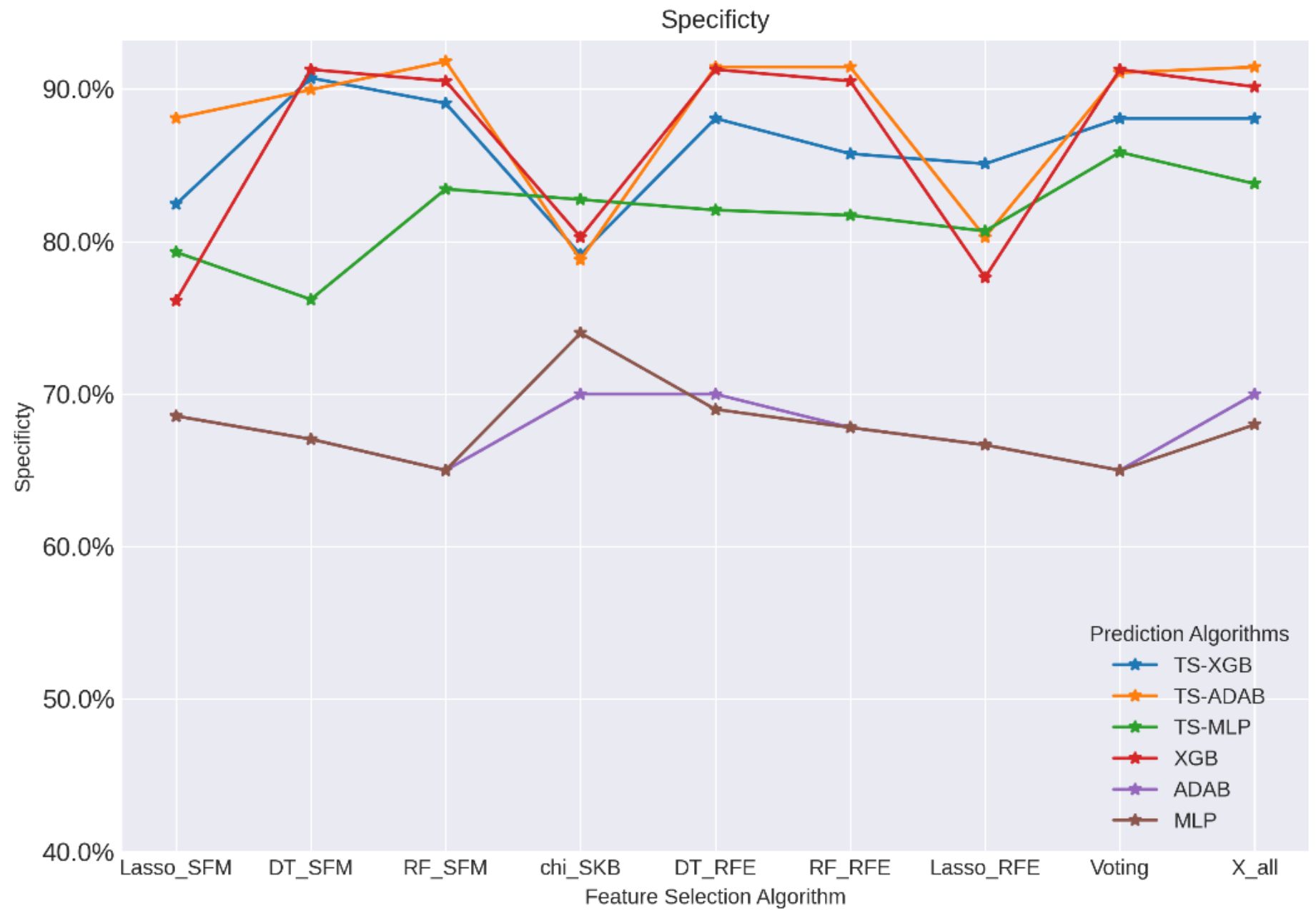

**Figure 13:** Specificity for all models.

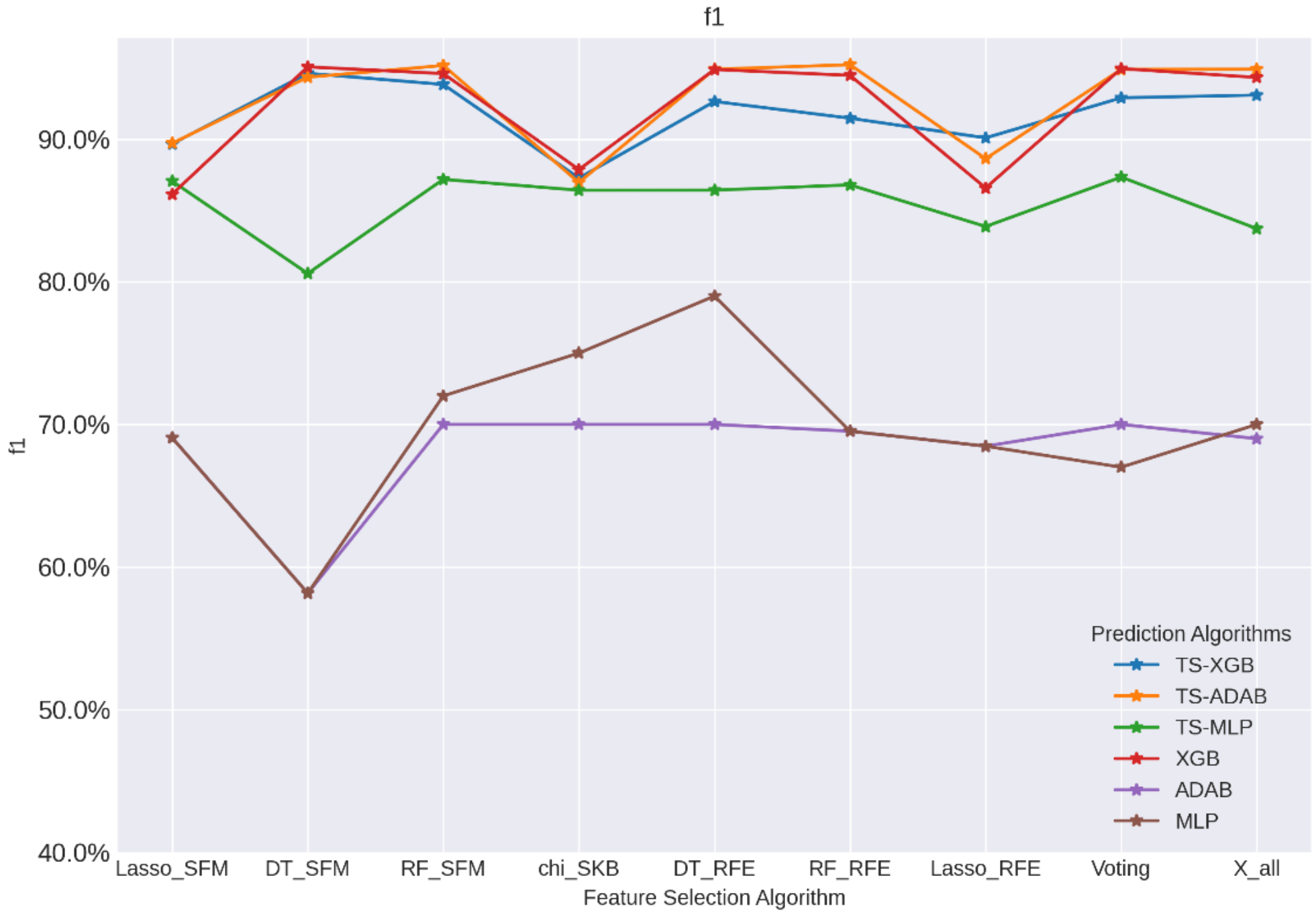

**Figure 14:** F1 for all models.

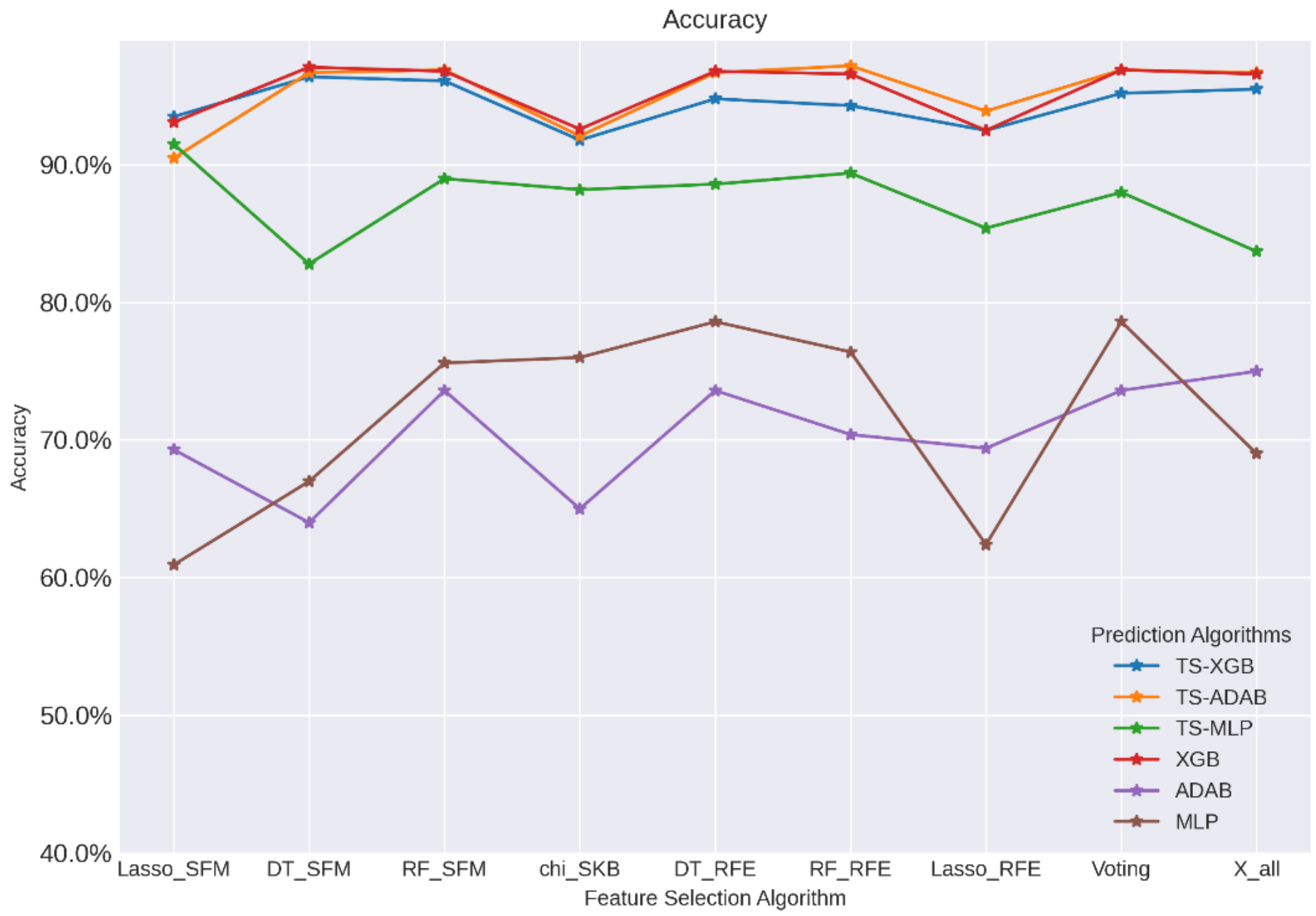

**Figure 15:** Accuracy for all models.

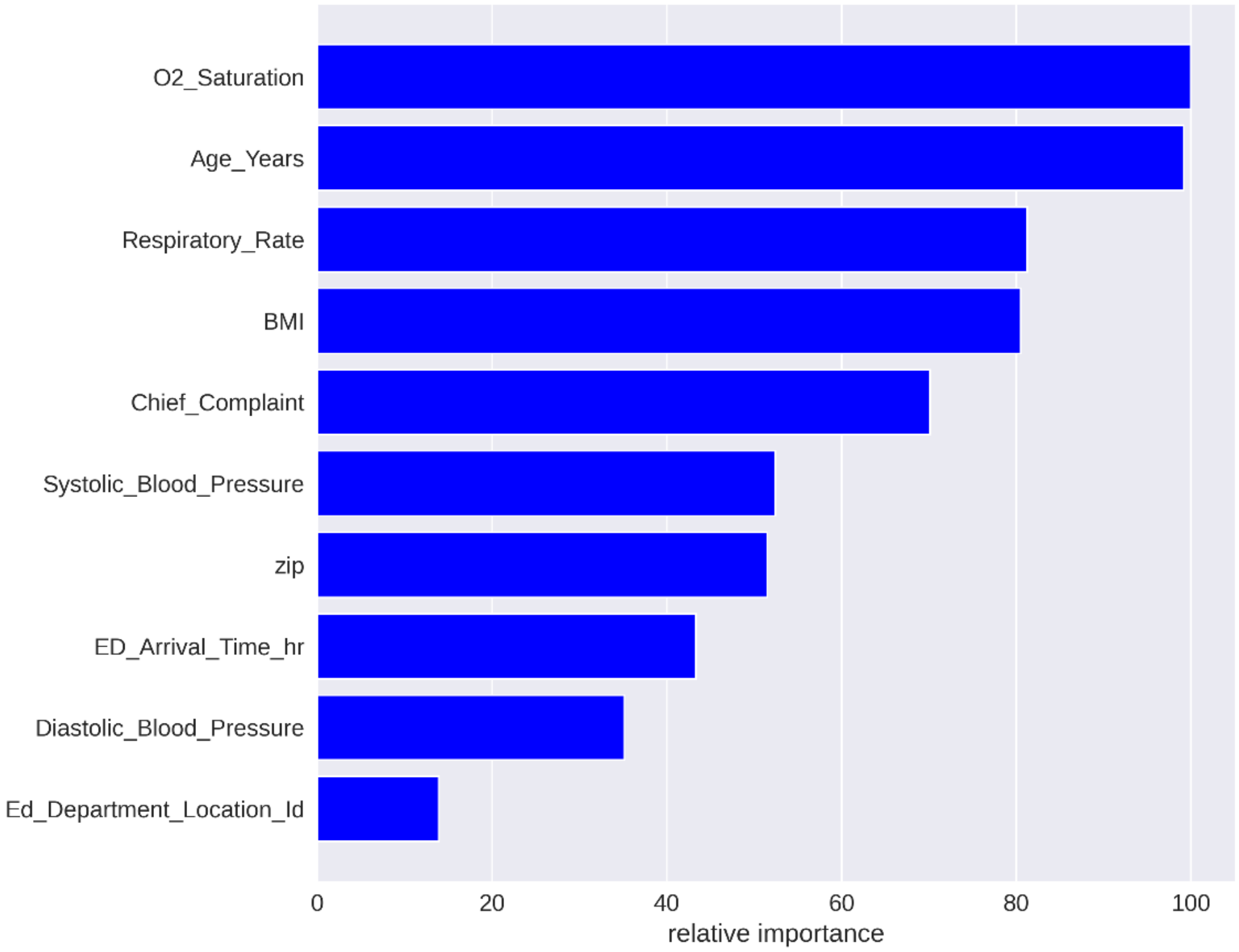

**Figure 16:** Feature importance based on the best model.

## 4.5 Sensitivity analysis

As mentioned in section 3.2, the sample size that is used for the model training is 5,000 out of 400k samples. The reason is that if the whole data is used, the computational cost of the three algorithms (e.g., T-XGB, T-ADAB, and T-MLP) will be very expensive. To confirm that the sample does not affect the quality of the proposed models, multiple experiments are conducted with different sample sizes on the best model. Figure 17 shows the AUCs for the best model with different sample sizes. It can be seen the accuracy of the model did not change significantly with increasing the sample size. Therefore, our model is robust and practical.

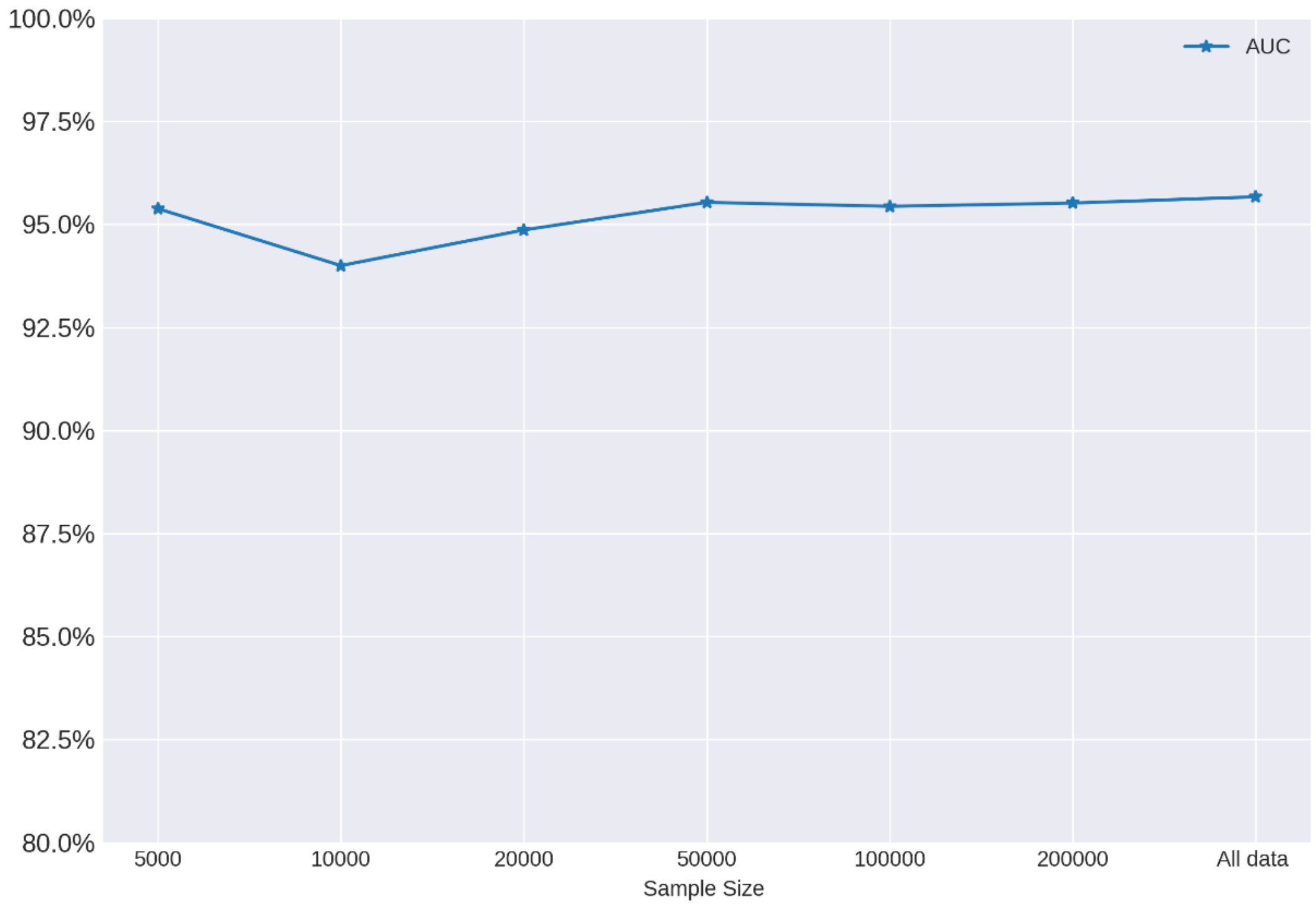

**Figure 17**: AUCs for the best model with different sample sizes.

# 5. MANAGERIAL IMPACT

Determining the admission status of an incoming patient to the ED is a very complex and dynamic decision. The complexity stems from the variety of conditions, illnesses, and injuries that healthcare providers encounter at EDs. The determination of the patient's severity along with deciding to admit a patient depends heavily on the healthcare providers' experience and patient's information at the time of triage. Delays or mistakes in making that decision could lead to long boarding times at EDs. Long boarding times are a crucial determinant of ED overcrowding. The proposed model helps in providing accurate and timely disposition decisions to efficiently manage the downstream operations in departments such as the inpatient unit. Since the final model includes vital signs and few demographic data, our proposed framework predict admission disposition decision at the moment (e.g., first five or ten minutes) an ED patient arrives in an ED. Efficient downstream operations such as bed assignment and transportation coordination are translated into shorter boarding times at EDs, thus patients move faster out of the ED with no delays. In this work, a decision model is developed to aid healthcare providers to forecast the admission status of ED patients in the early stages by using basic and readily available triage data. The tool is built by integrating optimization (e.g., Tabu search) and predictive algorithms. The use of this decision tool is expected to improve patient flow at ED, mitigate overcrowding and its related symptoms, and improve patient safety and healthcare quality.

## 6. CONCLUSION, LIMITATIONS, AND FUTURE WORK

This work expands the growing literature on the use of predictive analytics and machine learning models in healthcare. This paper focuses on developing a framework to create a practical decision tool derived from a prediction model to identify the admission status of patients at EDs. The accurate and early prediction of the admission status helps hospital administrators plan proactively the downstream operations which facilitate patient flow and improves resource planning and coordination across the hospital areas (e.g., ED and inpatient units). Efficient patient flow, resource management, and coordination can help to minimize patient boarding delays in EDs, thus reducing overcrowding.

In this work, retrospective patient data of all ED visits over three years to a large hospital has been used to build several machine learning models. The data include a total of 17 features per patient visit. The framework involved four feature selection algorithms including, decision tree (DT), Chi-square (Chi-sq), random forest (RF), and the least absolute shrinkage and selection operator logistic regression (Lasso-LR). Nine data groups result from the feature selection step. In addition, three new machine learning algorithms including T-XGB, T-MLP, and T-ADAB are used to build the prediction models. An intelligent metaheuristic algorithm (e.g., Tabu search) is used to optimize the machine learning algorithms parameters which resulted in a better performance. The three newly proposed algorithms are compared with the traditional ones (e.g., XGB, ADAB, and MLP). Based on the experiments performed in this study, the newly proposed algorithms (i.e., T-XGB, T-MLP, and T-ADAB) outperformed the traditional ones (i.e., XGB, MLP, and ADAB). T-ADAB algorithm, which is built using the RF_RFE data group, has provided the best results with the following performance measures: AUC (95.4%), sensitivity (99.3%), specificity (91.4%), F1 (95.2%), and accuracy (97. 2%). The features that resulted in the best model are O2 saturation, age, respiratory rate, BMI, chief complaint, systolic blood pressure, zip code, arrival time in hours, diastolic blood pressure, and ED location. The addition of the Tabu search (TS) algorithm to optimize the parameters of the machine learning algorithms has significantly improved the accuracy of the resulted models.

For the limitations of the proposed study, there are a few of them. Firstly, the models are trained using data from multiple locations of the partner hospital to obtain a more generalized model. However, this may negatively affect the performance of models at each specific location. Secondly, other factors can affect the admission disposition decision but not captured in this study such as patients' preferences with respect to hospitalization. To further the work presented in this paper, other patient statuses could be included such as left without being seen, expired patient, and transferred. Including such statuses could help hospitals better manage their resources. Furthermore, after implementing the models, investigating the ED operations including the boarding process at the partner hospital pre and post implementing of the proposed model can be a future. It will provide an insight into the effectiveness of machine learning in improving a hospital's operations.

**References**

Andradóttir, S. (2015). A review of random search methods. *Handbook of Simulation Optimization*, 277–292.

Araz, O. M., Olson, D., & Ramirez-Nafarrate, A. (2019). Predictive analytics for hospital admissions from the emergency department using triage information. *International Journal of Production Economics*, *208*, 199–207. https://doi.org/10.1016/j.ijpe.2018.11.024

Arya, R., Wei, G., McCoy, J. V., Crane, J., Ohman-Strickland, P., & Eisenstein, R. M. (2013). Decreasing length of stay in the emergency department with a split emergency severity index 3 patient flow model. *Academic Emergency Medicine*, *20*(11), 1171–1179. https://doi.org/10.1111/acem.12249

Ashour, O. M., & Okudan Kremer, G. E. (2013). A simulation analysis of the impact of FAHP-MAUT triage algorithm on the Emergency Department performance measures. *Expert Systems with Applications*, *40*(1), 177–187. https://doi.org/10.1016/j.eswa.2012.07.024

Ashour, O. M., & Okudan Kremer, G. E. (2016). Dynamic patient grouping and prioritization: A new approach to emergency department flow improvement. *Health Care Management Science*, *19*(2). https://doi.org/10.1007/s10729-014-9311-1

Bacchi, S., Gluck, S., Tan, Y., Chim, I., Cheng, J., Gilbert, T., Menon, D. K., Jannes, J., Kleinig, T., & Koblar, S. (2020). Prediction of general medical admission length of stay with natural language processing and deep learning: A pilot study. *Internal and Emergency Medicine*, *15*(6), 989–995.

Badrouchi, S., Ahmed, A., Mongi Bacha, M., Abderrahim, E., & Ben Abdallah, T. (2021). A machine learning framework for predicting long-term graft survival after kidney transplantation. *Expert Systems with Applications*, *182*, 115235. https://doi.org/10.1016/j.eswa.2021.115235

Barak-Corren, Y., Fine, A. M., & Reis, B. Y. (2017). Early prediction model of patient hospitalization from the pediatric emergency department. *Pediatrics*, *139*(5). https://doi.org/10.1542/peds.2016-2785

Barak-Corren, Y., Israelit, S. H., & Reis, B. Y. (2017). Progressive prediction of hospitalisation in the emergency department: Uncovering hidden patterns to improve patient flow. *Emergency Medicine Journal*, *34*(5), 308–314. https://doi.org/10.1136/emermed-2014-203819

Ben-Tovim, D. I., Bassham, J. E., Bennett, D. M., Dougherty, M. L., Martin, M. A., O'Neill, S. J., Sincock, J. L., & Szwarcbord, M. G. (2008). Redesigning care at the Flinders Medical Centre: Clinical process redesign using “lean thinking”. *The Medical Journal of Australia*, *188*(6 Suppl), 27–31. https://doi.org/10.5694/j.1326-5377.2008.tb01671.x

Bereta, M. (2019). Regularization of boosted decision stumps using tabu search. *Applied Soft Computing*, *79*, 424–438.

Bergstra, J., Bardenet, R., Bengio, Y., & Kégl, B. (2011). Algorithms for hyper-parameter optimization. *25th Annual Conference on Neural Information Processing Systems (NIPS 2011)*, *24*.

Bergstra, J., & Bengio, Y. (2012). Random search for hyper-parameter optimization. *Journal of Machine Learning Research*, *13*(2).

Bergstra, J., Yamins, D., & Cox, D. (2013). Making a science of model search: Hyperparameter optimization in hundreds of dimensions for vision architectures. *International Conference on Machine Learning*, 115–123.

Cameron, A., Jones, D., Logan, E., O’Keeffe, C. A., Mason, S. M., & Lowe, D. J. (2018). Comparison of Glasgow Admission Prediction Score and Amb Score in predicting need for inpatient care. *Emergency Medicine Journal*, *35*(4), 247–251. https://doi.org/10.1136/emermed-2017-207246

Chalfin, D. B., Trzeciak, S., Likourezos, A., Baumann, B. M., Dellinger, R. P., & Group, for the D.-E. study. (2007). Impact of delayed transfer of critically ill patients from the emergency department to the intensive care unit*. *Critical Care Medicine*, *35*(6), 1477–1483. https://doi.org/10.1097/01.CCM.0000266585.74905.5A

Chen, C., Zhang, Q., Yu, B., Yu, Z., Lawrence, P. J., Ma, Q., & Zhang, Y. (2020). Improving protein-protein interactions prediction accuracy using XGBoost feature selection and stacked ensemble classifier. *Computers in Biology and Medicine*, *123*, 103899. https://doi.org/10.1016/j.compbiomed.2020.103899

Chen, J., Zhao, F., Sun, Y., & Yin, Y. (2020). Improved XGBoost model based on genetic algorithm. *International Journal of Computer Applications in Technology*. https://www.inderscienceonline.com/doi/abs/10.1504/IJCAT.2020.106571

Chen, T., He, T., Benesty, M., Khotilovich, V., & Tang, Y. (2015). Xgboost: Extreme gradient boosting. *R Package Version 0.4-2*, 1–4.

Chonde, S. J., Ashour, O. M., Nembhard, D. A., & Kremer, G. E. O. (2013). Model comparison in Emergency Severity Index level prediction. *Expert Systems with Applications*, *40*(17), 6901–6909. https://doi.org/10.1016/j.eswa.2013.06.026

Chou, J.-S., Cheng, M.-Y., Wu, Y.-W., & Pham, A.-D. (2014). Optimizing parameters of support vector machine using fast messy genetic algorithm for dispute classification. *Expert Systems with Applications*, *41*(8), 3955–3964.

Considine, J., Kropman, M., Kelly, E., & Winter, C. (2008). Effect of emergency department fast track on emergency department length of stay: A case-control study. *Emergency Medicine Journal*, *25*(12), 815–819. https://doi.org/10.1136/emj.2008.057919

De Hond, A., Raven, W., Schinkelshoek, L., Gaakeer, M., Ter Avest, E., Sir, O., Lameijer, H., Hessels, R. A., Reijnen, R., De Jonge, E., Steyerberg, E., Nickel, C. H., & De Groot, B. (2021). Machine learning for developing a prediction model of hospital admission of emergency department patients: Hype or hope? *International Journal of Medical Informatics*, *152*, 104496. https://doi.org/10.1016/j.ijmedinf.2021.104496

Desautels, T., Calvert, J., Hoffman, J., Jay, M., Kerem, Y., Shieh, L., Shimabukuro, D., Chettipally, U., Feldman, M. D., Barton, C., Wales, D. J., & Das, R. (2016). Prediction of Sepsis in the Intensive Care Unit With Minimal Electronic Health Record Data: A Machine Learning Approach. *JMIR Medical Informatics*, *4*(3), e28. https://doi.org/10.2196/medinform.5909

Dickson, E. W., Singh, S., Cheung, D. S., Wyatt, C. C., & Nugent, A. S. (2009). Application of Lean Manufacturing Techniques in the Emergency Department. *Journal of Emergency Medicine*, *37*(2), 177–182. https://doi.org/10.1016/j.jemermed.2007.11.108

Dugas, A. F., Kirsch, T. D., Toerper, M., Korley, F., Yenokyan, G., France, D., Hager, D., & Levin, S. (2016). An Electronic Emergency Triage System to Improve Patient Distribution by Critical

Outcomes. *Journal of Emergency Medicine*, *50*(6), 910–918. https://doi.org/10.1016/j.jemermed.2016.02.026

El-Darzi, E., Abbi, R., Vasilakis, C., Gorunescu, F., Gorunescu, M., & Millard, P. (2009). Length of Stay-Based Clustering Methods for Patient Grouping. In S. Mcclean, P. Millard, E. El-darzi, & C. N. Eds (Eds.), *Intelligent Patient Management* (pp. 39–56). Springer-Verlag Berlin Heidelberg. https://doi.org/10.1109/cbms.2011.5999122

Fatovich, D. M., Nagree, Y., & Sprivulis, P. (2005). Access block causes emergency department overcrowding and ambulance diversion in Perth, Western Australia. *Emergency Medicine Journal*, *22*(5), 351–354. https://doi.org/10.1136/emj.2004.018002

Fernandes, M., Vieira, S. M., Leite, F., Palos, C., Finkelstein, S., & Sousa, J. M. (2020). Clinical decision support systems for triage in the emergency department using intelligent systems: A review. *Artificial Intelligence in Medicine*, *102*, 101762.

Futoma, J., Morris, J., & Lucas, J. (2015). A comparison of models for predicting early hospital readmissions. *Journal of Biomedical Informatics*, *56*, 229–238. https://doi.org/10.1016/j.jbi.2015.05.016

Gendreau, M., Potvin, J.-Y. (2019). *Handbook of Metaheuristics | Michel Gendreau | Springer*. https://www.springer.com/gp/book/9783319910857

Glover, F. (1986). Future paths for integer programming and links to artificial intelligence. Computers & operations research, 13(5), 533-549.

Golmohammadi, D. (2016). Predicting hospital admissions to reduce emergency department boarding. *International Journal of Production Economics*, *182*, 535–544.

Graham, B., Bond, R., Quinn, M., & Mulvenna, M. (2018). Using data mining to predict hospital admissions from the emergency department. *IEEE Access*, *6*, 10458–10469.

Guo, J., Yang, L., Bie, R., Yu, J., Gao, Y., Shen, Y., & Kos, A. (2019). An XGBoost-based physical fitness evaluation model using advanced feature selection and Bayesian hyper-parameter optimization for wearable running monitoring. *Computer Networks*, *151*, 166–180. https://doi.org/10.1016/j.comnet.2019.01.026

Haimovich, J. S., Venkatesh, A. K., Shojaee, A., Coppi, A., Warner, F., Li, S. X., & Krumholz, H. M. (2017). Discovery of temporal and disease association patterns in condition-specific hospital utilization rates. *PLoS ONE*, *12*(3), 1–15. https://doi.org/10.1371/journal.pone.0172049

Han, Jiawei., & Kamber, Micheline. (2001). *Data mining: Concepts and techniques*. Morgan Kaufmann Publishers.

Hastie, T., Tibshirani, R., & Friedman, J. (2009). *The elements of statistical learning: Data mining, inference, and prediction*.

Holden, R. J. (2011). Lean thinking in emergency departments: A critical review. *Annals of Emergency Medicine*, *57*(3), 265–278. https://doi.org/10.1016/j.annemergmed.2010.08.001

Hong, W. S., Haimovich, A. D., & Taylor, R. A. (2018). Predicting hospital admission at emergency department triage using machine learning. *PLoS ONE*, *13*(7), 1–13. https://doi.org/10.1371/journal.pone.0201016

Hoot, N. R., & Aronsky, D. (2008). Systematic Review of Emergency Department Crowding: Causes, Effects, and Solutions. *Annals of Emergency Medicine*, *52*(2). https://doi.org/10.1016/j.annemergmed.2008.03.014

Horng, S., Sontag, D. A., Halpern, Y., Jernite, Y., Shapiro, N. I., & Nathanson, L. A. (2017). Creating an automated trigger for sepsis clinical decision support at emergency department triage using machine learning. *PLoS ONE*, *12*(4), 1–16. https://doi.org/10.1371/journal.pone.0174708

Jamel, T. M., & Khammas, B. M. (2012). Implementation of a sigmoid activation function for neural network using FPGA. *13th Scientific Conference of Al-Ma'moon University College*, *13*.

Kelly, A. M., Bryant, M., Cox, L., & Jolley, D. (2007). Improving emergency department efficiency by patient streaming to outcomes-based teams. *Australian Health Review : A Publication of the Australian Hospital Association*, *31*(1), 16–21. https://doi.org/10.1071/ah070016

King, D. L., Ben-Tovim, D. I., & Bassham, J. (2006). Redesigning emergency department patient flows: Application of Lean Thinking to health care. *EMA - Emergency Medicine Australasia*, *18*(4), 391–397. https://doi.org/10.1111/j.1742-6723.2006.00872.x

Kohavi, R., & Quinlan, R. (2002). Data mining tasks and methods: Classification: Decision-tree discovery. *Handbook of Data Mining and Knowledge Discovery*, 267–276.

Kumar, A., & Jain, M. (2020). *Ensemble Learning for AI Developers*. Springer.

Lee, S. Y., Chinnam, R. B., Dalkiran, E., Krupp, S., & Nauss, M. (2020). Prediction of emergency department patient disposition decision for proactive resource allocation for admission. *Health Care Management Science*, *23*(3), 339–359. https://doi.org/10.1007/s10729-019-09496-y

Leegon, J., Jones, I., Lanaghan, K., & Aronsky, D. (2005). Predicting hospital admission for Emergency Department patients using a Bayesian network. *AMIA ... Annual Symposium Proceedings / AMIA Symposium. AMIA Symposium*, 1022.

Leegon, J., Jones, I., Lanaghan, K., & Aronsky, D. (2006). Predicting hospital admission in a pediatric Emergency Department using an Artificial Neural Network. *AMIA ... Annual Symposium Proceedings / AMIA Symposium. AMIA Symposium*, 1004.

Levin, S., Toerper, M., Hamrock, E., Hinson, J. S., Barnes, S., Gardner, H., Dugas, A., Linton, B., Kirsch, T., & Kelen, G. (2018). Machine-Learning-Based Electronic Triage More Accurately Differentiates Patients With Respect to Clinical Outcomes Compared With the Emergency Severity Index. *Annals of Emergency Medicine*, *71*(5), 565-574.e2. https://doi.org/10.1016/j.annemergmed.2017.08.005

Li, L., Jamieson, K., DeSalvo, G., Rostamizadeh, A., & Talwalkar, A. (2017). Hyperband: A novel bandit-based approach to hyperparameter optimization. *The Journal of Machine Learning Research*, *18*(1), 6765–6816.

Lucini, F. R., S. Fogliatto, F., Giovani, G. J., L. Neyeloff, J., Anzanello, M. J., de S. Kuchenbecker, R., & D. Schaan, B. (2017). Text mining approach to predict hospital admissions using early medical records from the emergency department. *International Journal of Medical Informatics*, *100,* 1–8. https://doi.org/10.1016/j.ijmedinf.2017.01.001

Lucke, J. A., De Gelder, J., Clarijs, F., Heringhaus, C., De Craen, A. J. M., Fogteloo, A. J., Blauw, G. J., De Groot, B., & Mooijaart, S. P. (2018). Early prediction of hospital admission for emergency department patients: A comparison between patients younger or older than 70 years. *Emergency Medicine Journal*, *35*(1), 18–27. https://doi.org/10.1136/emermed-2016-205846

Ma, L., Fu, T., Blaschke, T., Li, M., Tiede, D., Zhou, Z., Ma, X., & Chen, D. (2017). Geo-Information Evaluation of Feature Selection Methods for Object-Based Land Cover Mapping of Unmanned Aerial Vehicle Imagery Using Random Forest and Support Vector Machine Classifiers. *ISPRS Int. J. Geo-Inf, 6,* 51. https://doi.org/10.3390/ijgi6020051

Moons, K. G. M., Kengne, A. P., Woodward, M., Royston, P., Vergouwe, Y., Altman, D. G., & Grobbee, D. E. (2012). Risk prediction models: I. Development, internal validation, and assessing the incremental value of a new (bio)marker. *Heart*, *98*(9), 683–690. https://doi.org/10.1136/heartjnl-2011-301246

Moore, B. J., Ph, D., Stocks, C., Ph, D., Owens, P. L., & Ph, D. (2017). *Trends in Emergency Department Visits, 2006-2014. HCUP Statistical Brief #227*.

Niel, O., & Bastard, P. (2019). Artificial Intelligence in Nephrology: Core Concepts, Clinical Applications, and Perspectives. *American Journal of Kidney Diseases*, *74*(6), 803–810. https://doi.org/10.1053/j.ajkd.2019.05.020

Obermeyer, Ziad, M. D., & Emanuel, Ezekiel J., M.D., Ph. D. (2016). Predicting the Future—Big Data, Machine Learning, and Clinical Medicine. *New England Journal of Medicine*, *375*(13), 1212–1216. https://doi.org/10.1056/NEJMp1606181.Predicting

Parker, C. A., Liu, N., Wu, S. X., Shen, Y., Lam, S. S. W., & Ong, M. E. H. (2019). Predicting hospital admission at the emergency department triage: A novel prediction model. *American Journal of Emergency Medicine*, *37*(8), 1498–1504. https://doi.org/10.1016/j.ajem.2018.10.060

Peck, J. S., Benneyan, J. C., Nightingale, D. J., & Gaehde, S. A. (2012). Predicting emergency department inpatient admissions to improve same-day patient flow. *Academic Emergency Medicine*, *19*(9), E1045–E1054. https://doi.org/10.1111/j.1553-2712.2012.01435.x

Pham, H. N. A., & Triantaphyllou, E. (2011). A meta-heuristic approach for improving the accuracy in some classification algorithms. *Computers & Operations Research*, *38*(1), 174–189. https://doi.org/10.1016/j.cor.2010.04.011

Pines, J. M., & Bernstein, S. L. (2015). Solving the worldwide emergency department crowding problem—What can we learn from an Israeli ED? *Israel Journal of Health Policy Research*, *4*(1), 10–13. https://doi.org/10.1186/s13584-015-0049-0

Pines, J. M., Hilton, J. A., Weber, E. J., Alkemade, A. J., Al Shabanah, H., Anderson, P. D., Bernhard, M., Bertini, A., Gries, A., Ferrandiz, S., Kumar, V. A., Harjola, V. P., Hogan, B., Madsen, B., Mason, S., Öhlén, G., Rainer, T., Rathlev, N., Revue, E., … Schull, M. J. (2011). International perspectives on emergency department crowding. *Academic Emergency Medicine*, *18*(12), 1358–1370. https://doi.org/10.1111/j.1553-2712.2011.01235.x

Putatunda, S., & Rama, K. (2019). A Modified Bayesian Optimization based Hyper-Parameter Tuning Approach for Extreme Gradient Boosting. *2019 Fifteenth International Conference on Information Processing (ICINPRO)*, 1–6. https://doi.org/10.1109/ICInPro47689.2019.9092025

Qiu, S., Chinnam, R. B., Murat, A., Batarse, B., Neemuchwala, H., & Jordan, W. (2015). A cost sensitive inpatient bed reservation approach to reduce emergency department boarding times. *Health Care Management Science*, *18*(1), 67–85. https://doi.org/10.1007/s10729-014-9283-1

Rätsch, G., Onoda, T., & Müller, K.-R. (2001). Soft margins for AdaBoost. *Machine Learning*, *42*(3), 287–320.

Raza, M. S., & Qamar, U. (2019). *Understanding and Using Rough Set Based Feature Selection: Concepts, Techniques and Applications*. Springer Nature.

Rodi, S. W., Grau, M. V., & Orsini, C. M. (2006). Evaluation of a fast track unit: Alignment of resources and demand results in improved satisfaction and decreased length of stay for emergency department patients. *Quality Management in Health Care*, *15*(3), 163–170. https://doi.org/10.1097/00019514-200607000-00006

Rui, P., & Kang K. (2017). National Hospital Ambulatory Medical Care Survey: 2017 emergency department summary tables. *National Ambulatory Medical Care Survey*, 37.

Saba, T. (2020). Recent advancement in cancer detection using machine learning: Systematic survey of decades, comparisons and challenges. *Journal of Infection and Public Health*, *13*(9), 1274–1289.

Sarkar, S., Vinay, S., Raj, R., Maiti, J., & Mitra, P. (2019). Application of optimized machine learning techniques for prediction of occupational accidents. *Computers & Operations Research*, *106*, 210–224.

Shi, P., Chou, M. C., Dai, J. G., Ding, D., & Sim, J. (2016). Models and insights for hospital inpatient operations: Time-dependent ED boarding time. *Management Science*, *62*(1), 1–28. https://doi.org/10.1287/mnsc.2014.2112

*sklearn.neural_network.MLPClassifier—Scikit-learn 0.24.0 documentation*. Retrieved December 30, 2020, from https://scikit-learn.org/stable/modules/generated/sklearn.neural_network.MLPClassifier.html

Sugumaran, V., Muralidharan, V., & Ramachandran, K. I. (2007). Feature selection using Decision Tree and classification through Proximal Support Vector Machine for fault diagnostics of roller bearing. *Mechanical Systems and Signal Processing*, *21*(2), 930–942. https://doi.org/10.1016/j.ymssp.2006.05.004

Sun, B. C., Hsia, R. Y., Weiss, R. E., Zingmond, D., Liang, L. J., Han, W., McCreath, H., & Asch, S. M. (2013). Effect of emergency department crowding on outcomes of admitted patients. *Annals of Emergency Medicine*, *61*(6), 605-611.e6. https://doi.org/10.1016/j.annemergmed.2012.10.026

Sun, Y., Heng, B. H., Tay, S. Y., & Seow, E. (2011). Predicting hospital admissions at emergency department triage using routine administrative data. *Academic Emergency Medicine*, *18*(8), 844–850. https://doi.org/10.1111/j.1553-2712.2011.01125.x

Tanabe, P., Gimbel, R., Yarnold, P. R., & Adams, J. G. (2004). The Emergency Severity Index (version 3) 5-level triage system scores predict ED resource consumption. *Journal of Emergency Nursing*, *30*(1), 22–29. https://doi.org/10.1016/j.jen.2003.11.004

Taylor, R. A., Pare, J. R., Venkatesh, A. K., Mowafi, H., Melnick, E. R., Fleischman, W., & Hall, M. K. (2016). Prediction of In-hospital Mortality in Emergency Department Patients with Sepsis: A Local Big Data-Driven, Machine Learning Approach. *Academic Emergency Medicine*, *23*(3), 269–278. https://doi.org/10.1111/acem.12876

Tsai, C.-C., & Li, S. H. (2009). A two-stage modeling with genetic algorithms for the nurse scheduling problem. *Expert Systems with Applications*, *36*(5), 9506–9512.

van der Vaart, T., Vastag, G., & Wijngaard, J. (2011). Facets of operational performance in an emergency room (ER). *International Journal of Production Economics*, *133*(1), 201–211. https://doi.org/10.1016/j.ijpe.2010.04.023

Verikas, A., Gelzinis, A., recognition, M. B.-P., & 2011, undefined. (n.d.). Mining data with random forests: A survey and results of new tests. *Elsevier*.

Weng, S. F., Reps, J., Kai, J., Garibaldi, J. M., & Qureshi, N. (2017). Can Machine-learning improve cardiovascular risk prediction using routine clinical data? *PLoS ONE*, *12*(4), 1–14. https://doi.org/10.1371/journal.pone.0174944

*XGBoost Parameters—Xgboost 1.4.0-SNAPSHOT documentation*. (n.d.). Retrieved February 28, 2021, from https://xgboost.readthedocs.io/en/latest/parameter.html#parameters-for-tree-booster

Xu, M., Watanachaturaporn, P., Varshney, P. K., & Arora, M. K. (2005). Decision tree regression for soft classification of remote sensing data. *Remote Sensing of Environment*, *97*(3), 322–336. https://doi.org/10.1016/j.rse.2005.05.008

Yu, C.-S., Lin, Y.-J., Lin, C.-H., Lin, S.-Y., Wu, J. L., & Chang, S.-S. (2020). Development of an Online Health Care Assessment for Preventive Medicine: A Machine Learning Approach. *Journal of Medical Internet Research*, *22*(6), e18585. https://doi.org/10.2196/18585

Yu, D., Liu, Z., Su, C., Han, Y., Duan, X., Zhang, R., Liu, X., Yang, Y., & Xu, S. (2020). Copy number variation in plasma as a tool for lung cancer prediction using Extreme Gradient Boosting (XGBoost) classifier. *Thoracic Cancer*, *11*(1), 95–102. https://doi.org/10.1111/1759-7714.13204